\documentclass[10pt,twocolumn,letterpaper]{article}

\usepackage[pagenumbers]{arxiv} %

\usepackage{graphicx}
\usepackage{amsmath}
\usepackage{amssymb}
\usepackage{booktabs}
\usepackage{glossaries}

\usepackage[pagebackref,breaklinks,colorlinks]{hyperref}

\usepackage[capitalize]{cleveref}
\crefname{section}{Sec.}{Secs.}
\Crefname{section}{Section}{Sections}
\Crefname{table}{Table}{Tables}
\crefname{table}{Tab.}{Tabs.}

\glsdisablehyper

\newacronym{ai}{AI}{Artificial Intelligence}
\newacronym{dl}{DL}{Deep Learning}
\newacronym{dnn}{DNN}{Deep Neural Network}
\newacronym{lrp}{LRP}{Layer-wise Relevance Propagation}
\newacronym{xai}{XAI}{eXplainable Artificial Intelligence}
\newacronym{crp}{CRP}{Concept Relevance Propagation}
\newacronym{amax}{ActMax}{Activation Maximization}
\newacronym{rmax}{RelMax}{Relevance Maximization}
\newacronym{auc}{AUC}{Area Under Curve}
\newacronym{aoc}{AOC}{Area Over Curve}
\newacronym{roi}{ROI}{Region of Interest}
\newacronym{sem}{SEM}{Standard Error of Mean}
\newacronym{lcrp}{L-CRP}{CRP for Localization Models}

\newcommand{\x}{{\mathbf{x}}\xspace}
\newcommand{\argmax}{{\text{argmax}}\xspace} 
\newcommand{\preact}{z}

\begin{document} 

\title{Revealing Hidden Context Bias in Segmentation and Object Detection through Concept-specific Explanations}

\author{Maximilian Dreyer$^{1}$ \and
Reduan Achtibat$^1$ \and
Thomas Wiegand$^{1,2,3}$ \and
Wojciech Samek$^{1,2,3,\dagger}$ \and
Sebastian Lapuschkin$^{1,\dagger}$}

\date{
\footnotesize
$^1$ Fraunhofer Heinrich-Hertz-Institute, 10587 Berlin, Germany\\
$^2$ Technische Universität Berlin, 10587 Berlin, Germany\\
$^3$ BIFOLD – Berlin Institute for the Foundations of Learning and Data, 10587 Berlin, Germany\\
$^\dagger$ corresponding authors: \texttt{\{wojciech.samek,sebastian.lapuschkin\}@hhi.fraunhofer.de}
} 

\maketitle

\begin{abstract}

Applying traditional post-hoc attribution methods to segmentation or object detection predictors offers only limited insights, as the obtained feature attribution maps at input level typically resemble the models' predicted segmentation mask or bounding box.
In this work, we address the need for more informative explanations for these predictors by proposing the post-hoc \glsdesc{xai} method L-CRP
to generate explanations that automatically identify and visualize relevant concepts learned, recognized and used by the model during inference as well as precisely locate them in input space. 
Our method therefore goes beyond singular input-level attribution maps and, as an approach based on the recently published \glsdesc{crp} technique, is efficiently applicable to state-of-the-art black-box architectures in segmentation and object detection,
such as DeepLabV3+ and YOLOv6, among others.
We verify the faithfulness of our proposed technique by quantitatively comparing different concept attribution methods, and discuss the effect on explanation complexity on popular datasets such as CityScapes, Pascal VOC and MS COCO 2017.
The ability to precisely locate and communicate concepts is used to reveal and verify the use of background features, thereby highlighting possible biases of the model.

\end{abstract}

\section{Introduction}

    \begin{figure*}  
                    \centering
                    \includegraphics[width=0.9\linewidth]{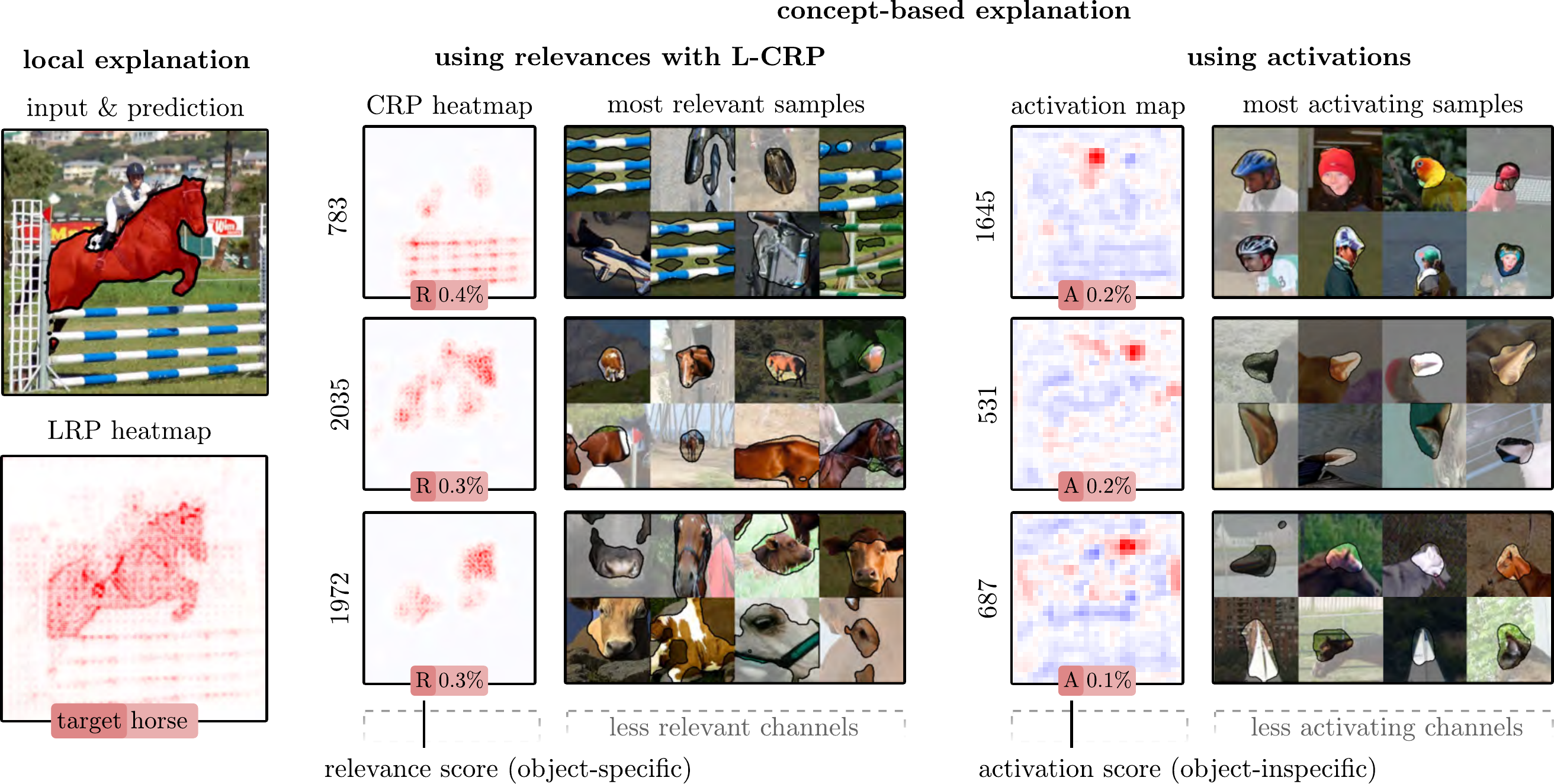}
                    \captionof{figure}{Our concept-based \gls{xai} approach (\emph{center}) goes beyond traditional local heatmaps (\emph{left}) or latent activation analysis (\emph{right}),
                    by communicating \emph{which} latent concepts contributed to a particular detection, \emph{to what degree} in terms of relevance (R) and \emph{where} concepts are precisely located in input space.
                    Whereas the shown heatmap from \gls{lrp} 
                    only indicates the relevance of pixels overall, 
                    our method disentangles the latent space and makes it possible to understand individual concepts forming the prediction outcome. 
                    Especially in the multi-object context,
                    it is crucial to attain object-class-specific explanations,
                    which is not possible by the analysis of latent activations:
                    Concepts with the highest activation values can refer to \emph{any} class that is present in the image,
                    such as the horse's rider, or none at all, since activations do not indicate whether a feature is actually used for inference. %
                    }
                    \label{fig:overview}
    \end{figure*}

\glspl{dnn} have proven to be successful in providing accurate predictions in several critical object localization tasks, including autonomous driving \cite{arnold2019survey} or medical screening \cite{mintz2019introduction}. 
However, 
the reasoning of these highly complex and non-linear models is generally not transparent \cite{rudin2019stop,samek2021explaining}, 
and as such, 
their decisions may be biased towards unintended or undesired features \cite{stock2018convnets,lapuschkin2019unmasking, anders2022finding}. 
In particular, object localization models sometimes base their decisions on features that lie outside of the segmentation mask or bounding box \cite{hoyer2019grid,petsiuk2021black}, as shown in Figure~\ref{fig:overview},
where the context of a hurdle is used for the segmentation of a horse.
Understanding such contextual usage by \glspl{dnn} is crucial to meet the requirements set in governmental regulatory frameworks and guidelines~\cite{goodman2017european, eu2019building}.

In order to increase our understanding of \gls{dnn} predictions,
the field of \glsfirst{xai}
has proposed several techniques that can be characterized as local, global or a combination of both.

In literature predominantly addressed are \emph{local} \gls{xai} methods for explaining single predictions.
These methods compute attribution scores of input features,
which can be visualized in form of heatmaps highlighting important input regions. 
However,
in object localization or segmentation tasks,
such attribution maps often resemble the prediction, \eg the segmentation mask itself, and hence add little value for gaining new insights into the decision process of a model, \eg when no context is used (see Figure~\ref{fig:local_to_glocal_seg} (\emph{right})).

Alternatively, 
\emph{global} \gls{xai} aims to visualize which features or concepts have been learned by a model or play an important role in a model’s reasoning in general.
Nonetheless, it is not clear which features were actually used for a particular prediction or how they interact.
For individual samples,
latent activation maps of features can be visualized, that however are of low resolution and give no indication whether the feature was actually used or merely present in the input, as illustrated in Figure~\ref{fig:overview} (\emph{right}).

\emph{Glocal} concept-based explanations,
on the other hand,
offer a new dimension of model understanding compared to traditional heatmap-based approaches by not only indicating \emph{where} the model pays attention to,
but also informing about \emph{what} it sees in the relevant input regions \cite{achtibat2022towards,olah2018building,zhou2018Interpretable}.
As far as we know,
we are hereby the first work to address glocal concept-based explantions for object localization tasks.

Our glocal methodology is hereby based on \gls{crp}~\cite{achtibat2022towards}, 
an extension to local attribution methods based on the (modified) model gradient, such as \glsfirst{lrp}~\cite{bach2015pixel}.
Specifically,
\gls{crp} allows to disentangle local attributions by computing relevance scores and heatmaps for individual latent features,
thereby making it possible to localize concepts precisely in input space.
Concept localization allows us in Section~\ref{sec:context} to identify context biases and the use of background features,
such as the ``vertical bar" concept targeting the hurdle in Figure~\ref{fig:overview}.

\paragraph{Contributions}
In this work,
we address the limited explainability insight of heatmap-based local \gls{xai} methods for object localization tasks. 
We therefore present a novel method called \gls{lcrp} for \emph{glocal} concept-based explanations based on \gls{crp} for state-of-the-art architectures for semantic segmentation and object detection tasks, including UNet~\cite{falk2019u}, DeepLabV3+~\cite{chen2018encoder}, YoloV5~\cite{jocher2021yolov5} and YoloV6~\cite{li2022yolov6} trained on the public datasets of CityScapes~\cite{cordts2016cityscapes}, Pascal VOC 2012~\cite{everingham2010pascal} and MS COCO 2017~\cite{lin2014microsoft}. 
Concretely,
\begin{enumerate}
    \item We demonstrate how to explain state-of-the-art architectures for segmentation as well as object detection with \gls{lrp} locally and by adapting \gls{crp} glocally on a concept level.

    \item We evaluate our method in terms of faithfulness and explanation complexity using
     different concept attribution methods, including activation, gradient, as well as several parameterizations of \gls{lrp}-attributions.

    \item Being able to localize concepts in input space via \gls{crp} heatmaps and having object masks or bounding boxes available, we compute context scores for concepts indicating to which degree a concept is used for encoding background or object features and how the model makes use of the respective information during inference. We show how these insights can be used to detect biases in the model and verify our findings by interacting with the model.
    
\end{enumerate}

\section{Related Work}
    In the following,
    the current landscape of \gls{xai} methods for segmentation and object detection models is presented,
    followed by a description of concept-based \gls{xai} techniques.
    
    \subsection{Explainable AI for Segmentation}
        
        The literature of \gls{xai} for segmentation predominantly focuses on local techniques.
        While some works apply backpropagation-based methods such as Grad-CAM \cite{vinogradova2020towards,mankodiya2022od,evans2022explainability} or \gls{lrp} \cite{ahmed2021explainable} to compute attribution heatmaps,
        others propose to apply perturbation-based  methods \cite{wan2020segnbdt,hoyer2019grid}. 
        Contrary to classification tasks, 
        local attributions are not computed \wrt a single output class neuron, but \wrt the whole output feature map that forms the segmentation mask representing the class.
        Several such local \gls{xai} techniques have been compiled into the Neuroscope toolbox \cite{schorr2021neuroscope}. 

        Alternatively, 
        Losch et al. inspect latent features of segmentation models and introduce Semantic BottleNecks \cite{losch2021semantic} modules to increase the latent space's human interpretability.
        They visualize and investigate intermediate filter activation maps, which, however, are of limited insight for understanding particular predictions due to not being class- and outcome-specific.
        Further,
        the fidelity of activation maps is limited to a convolutional channel's spatial resolution, as shown in Figure~\ref{fig:overview},
        where the attributions from \gls{lcrp} offer high resolution, as well as object- and outcome-specificity.

        Alternatively,
        another group of works investigates inherently interpretable architectures,
        such as U-Noise~\cite{koker2021u}, SegNBDT \cite{wan2020segnbdt} or MSGA-Net \cite{karri2022explainable}.
        However,
        a large number of models applied in industry and research are not designed to be human-interpretable in the first place and thus require post-hoc methods for interpretation, such as ours.
        
    \subsection{Explainable AI for Object Detection}
        Similar to image segmentation,
        several local \gls{xai} methods have been presented for object detection.
        These techniques can be grouped into methods based on the (modified) gradient such as Gradient-SHAP \cite{kawauchi2022shap},
        LRP \cite{tsunakawa2019contrastive,karasmanoglou2022heatmap}, 
        Spatial Sensitive Grad-CAM \cite{yamauchi2022spatial} and EX2 \cite{gudovskiy2018explain},
        or input-perturbation techniques such as LIME \cite{gudovskiy2018explain} and masking \cite{petsiuk2021black,yan2022gsm,schinagl2022occam}.
        Methods based on the gradient hereby explain the output class logit
        of a chosen bounding box
        analogously to the classification case.
        
        It is to note,
        that perturbation-based attribution methods require a high number of prediction re-evaluations by the model,
        resulting in run times in the order of minutes per data point, as \eg for \cite{petsiuk2021black}.
        Our method is based on the (modified) gradient and, therefore, can be computed in the order of seconds \cite{achtibat2022towards}.

    \subsection{Concept-based Explanations}

        In recent years, a multitude of methods emerged to visualize in a human-interpretable way concepts in the latent space learned by a model.
        A line of work~\cite{bau2020understanding, cammarata2020thread} assumes that individual neurons encode distinct concepts,
        others view concepts as directions described by a superposition of neurons \cite{kim2018interpretability, vielhaben2022sparse}.
        Similar to contemporary literature~\cite{achtibat2022towards, bau2020understanding, cammarata2020thread}, we treat each neuron or convolutional feature map as an independent concept to achieve the highest granularity in explanations, while the method presented in this paper can be, in principle, also extended trivially to concept directions in the latent space.
        
        For models operating in the image domain, contemporary work relies on activation maximization for visualizing concepts~\cite{zhou2015object, olah2017feature, radford2017learning, goh2021multimodal}, where in its simplest form, input images are sought that give rise to the highest activation value of a specific concept unit. 
        However, 
        high activation does not necessitate that corresponding input features are representative of a neuron’s function, as adversarial examples illustrate.
        In this work, 
        we make use of \gls{rmax}~\cite{achtibat2022towards} that mitigates the aforementioned issues by choosing reference images from the original training distribution based on maximal relevance instead of activation. 
        
        Glocal \gls{xai} methods try to bridge the gap between the visualization of concepts on a global scale and attribution of their role during per-sample model inference. 
        Shrouff et al. ~\cite{schrouff2021best} combine TCAV~\cite{kim2018interpretability} with Integrated Gradients to enable local attributions of neuron vectors,
        however, without offering localization of latent features in input space. 
        Achtibat et al. \cite{achtibat2022towards} propose the idea of \gls{crp}, an extension of latent feature attribution methods based on the (modified) gradient,
        where a concept-specific heatmap can be computed by restricting the backward propagation of attributions through the network.
        This allows to attain precise concept-conditional heatmaps in the input space,
        which we will use in Section \ref{sec:context} to investigate the use of background features by the model.
    
\section{Methods}
    Our glocal concept-based method \gls{lcrp} is based on the principle of \gls{crp},
    with \gls{lrp} as the feature attribution method of our choice.
    Therefore,
    we first introduce \gls{lrp} and \gls{crp} to attain concept-based explanations for individual predictions.
    Thereafter,
    we describe how concept-based explanations with \gls{lcrp} for segmentation and object detection can be obtained.

    \subsection{Layerwise-Relevance-Propagation}
    \label{sec:methods:lrp}

        \glsdesc{lrp} \cite{bach2015pixel} is an attribution method
        based on the conservation of flows and proportional decomposition.
        For a model $f(x)=f_n \circ \dots \circ  f_1(x)$ with $n$ layers,
        \gls{lrp} first calculates all activations during the forward pass starting with $f_1$ until the output layer $f_n$ is reached. 
        Thereafter, 
        the prediction score $f(x)$ of any chosen model output class is redistributed  as an initial quantity of relevance $R_n$ back towards the input layer after layer. 
        
        Given a layer's output neuron $j$, the distribution of its assigned relevance score $R_j$ towards its lower layer input neurons $i$ can be, in general, achieved by applying the basic decomposition rule
        \begin{equation} \label{eq:lrp_basic_decomp}
            R_{i \leftarrow j} = \frac{z_{ij}}{z_j}R_j~,
        \end{equation}
        with $z_{ij}$ describing the contribution of neuron $i$ to the activation of neuron $j$. %
        The aggregated pre-activations $z_{ij}$ at output neuron $j$ are represented by $z_j$ with $z_j = \sum_i z_{ij}$.
        The relevance of a neuron $i$ is then simply an aggregation of all incoming relevance quantities
        \begin{equation} \label{eq:lrp_basic_aggregate}
            R_i = \sum_j R_{i \leftarrow j}~.
        \end{equation}

        In order to ensure robust decompositions and stable heatmaps, 
        several purposed \gls{lrp} rules have been introduced in literature \cite{montavon2019layer,samek2021explaining}. 
        During the experiments in Section~\ref{sec:experiments}, 
        we compare the rules of \gls{lrp}-$\varepsilon$, \gls{lrp}-$\gamma$ and \gls{lrp}-z$^+$,
        leading to different explanations in terms of faithfulness and complexity
        as
        shown in Section~\ref{sec:explanation_quality}.
        Please refer to the Appendix~\ref{app:technical_details} for a detailed description of used \gls{lrp}-rules and how they are applied to the models.%
        
    \subsection{Concept-Relevance-Propagation}
        With \gls{crp}, 
        the authors of \cite{achtibat2022towards} combine global concept visualization techniques with the local feature attribution method of \gls{lrp}.
        A first step to the unification of both local and global \gls{xai} is the realization
        that during the \gls{lrp} backward pass, intermediate relevance scores are readily available, 
        as computed in Equation~\eqref{eq:lrp_basic_aggregate}.
        However,
        in order to also achieve concept-conditional heatmaps in input space,
        \gls{crp} firstly proposes to restrict the relevance propagation process of \gls{lrp} via conditions. 
        
        Concretely,
        a condition $c_l$ can be specified for one or multiple neurons $j$ corresponding to concepts of interest in a layer $l$.
        Multiple such conditions are combined to a condition set $\theta$.
        To disentangle attributions for latent representations, the relevance decomposition formula in Equation~\eqref{eq:lrp_basic_aggregate} is therefore extended with a ``filtering'' functionality:
        \begin{align}
            R^{(l-1,l)}_{i \leftarrow j}(\x|\theta \cup \theta_{l}) = \frac{z_{ij}}{z_j} \cdot \sum_{c_l \in \theta_l}\delta_{jc_l} \cdot R^l_j(\x|\theta) 
            \label{eq:appendix:lrp:masked}
        \end{align}
        where $\delta_{jc_l}$ ``selects'' the relevance quantity $R^l_j$ of layer $l$ and neuron $j$ for further propagation, if $j$ meets the (identity) condition(s) $c_l$ tied to concepts we are interested in. 
        For layers without conditions, no filtering is applied.
        A concept-conditional heatmap can thus be computed by conditioning the modified backward pass of \gls{lrp}
        via such a condition $c_l$ for the concept's corresponding neuron or filter.
        
        For the visualization of concepts,
        we adhere to the proposition of \cite{achtibat2022towards} and collect reference input images for which a latent neuron is most relevant, \ie, useful during inference.
        Thereafter,
        by computing a conditional heatmap for the neuron of interest and reference sample,
        the relevant input part is further cropped out and masked
        to increase the focus of the given explanation on the core features encoded by the investigated neurons,
        as detailed in \cite{achtibat2022towards}.
    
    \subsection{Extending Attributions to Object Localization}
    
        In order to obtain \gls{lcrp} for the generation of local explanations for segmentation and object detection models with \gls{crp},
        the task-specific output vectors and maps have to be handled accordingly. 
    
        \paragraph{Semantic Segmentation}
        
            Image segmentation models follow an encoder-decoder architecture in which the output mirrors (a scaled version of) the input dimensions in width and height. 
            In contrast to the 1-dimensional output in classification tasks, 
            the segmentation output consists of a 2d-map for each learned object category. 
            
            For simplicity, we will assume a decision function $\mathbf f$ for binary segmentation with a one dimensional input $\x \in \mathbb{R}^n$ and output $\mathbf f(\x) \in \mathbb{R}^n$ of length $n$ in the following, 
            \eg,
            $f: \mathbb{R}^{n} \rightarrow \mathbb{R}^{n}$. 
            This is not necessarily a restriction to the input and output size, 
            as any input can be flattened to a single dimension, \eg, $n = whc$ for images of size $w\times h\times c$. 
            
            Here,
            any input feature $x_i$ (\eg pixel)
            might contribute to all of the output values $f_j(\x)$.
            Therefore,
            we can assign to feature $x_i$ a relevance score for each output $j$, 
            that we summarize in a relevance vector $\mathbf{R}_i(\x|\theta)\in \mathbb{R}^n$.
            In fact,
            explaining a segmentation prediction can also be viewed as performing an explanation for each output pixel separately,
            and eventually adding the resulting attribution scores via, \eg, a weighted sum, as

            \begin{equation}
                R_i (\x|\theta)
                = \sum_j w_{j} \left(\mathbf R_i(\x|\theta)\right)_j 
                \label{eq:methods:lrp:attribution-vector:segmentation}
            \end{equation}
            with weights $w_{i} \in \mathbb{R}$.

            \begin{figure}  
                \centering
                \includegraphics[width=1\linewidth]{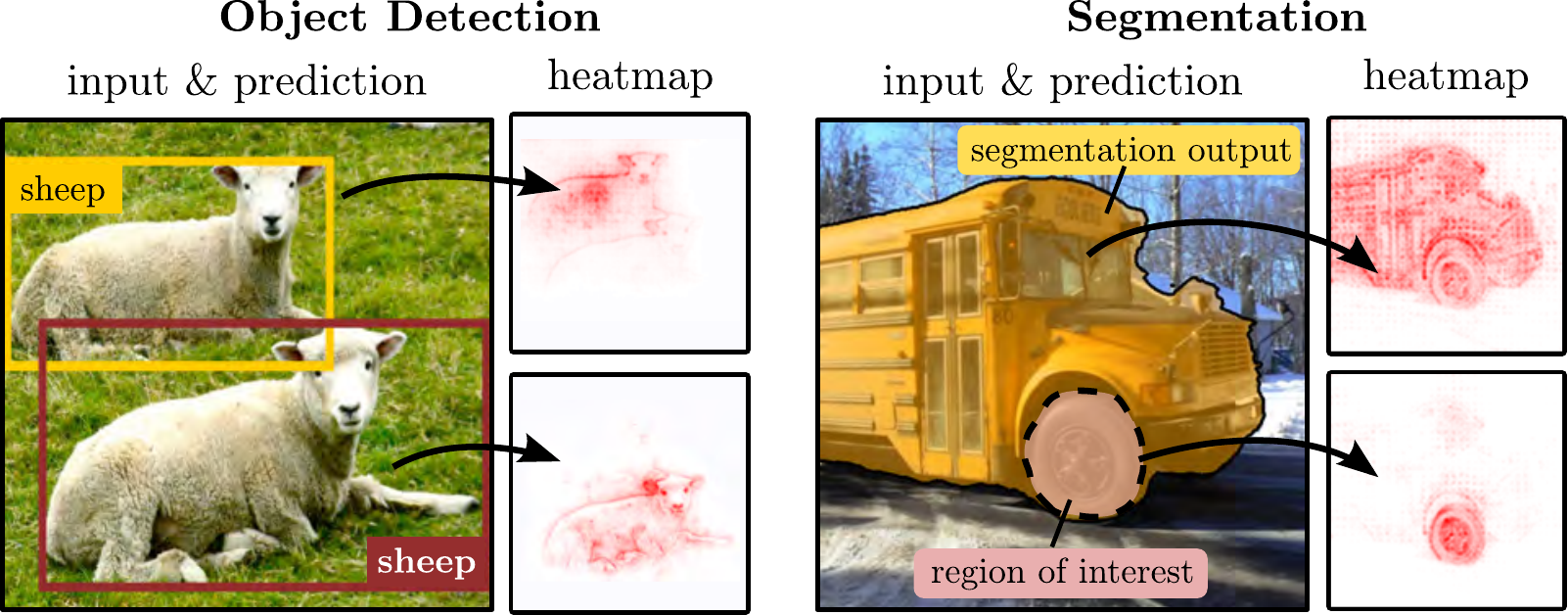}
                \captionof{figure}{Local explanations of object detection (\emph{left}) and semantic segmentation (\emph{right}) with the \gls{lrp}-z$^+$ rule. In the object detection case,
                we can explain each individual predicted object. 
                Similarly,
                for semantic segmentation,
                the output segmentation map of an object of choice can be explained.
                Here, in principle,
                also parts of the segmentation can be investigated.
                Concept-based explanation variants with \gls{lcrp} can be seen for both object detection and segmentation in Appendix~\ref{app:examples}.}
                \label{fig:local_to_glocal_seg}
            \end{figure} 
            
            The weights can be used to control the meaning and outcome of the explanation.
            Here,
            a specific \gls{roi} can be selected to be explained by setting $w_{i}=0$ for pixels $i$ outside the \gls{roi}. 
            Further, 
            the relevance values from different output pixels can be weighted differently.
            Choices include to weight scores according to their model confidence or,
            equally,
            by propagation of ones.
            For modified backpropagation-based attribution methods such as \gls{lrp},
            the output tensor used as the starting point for relevance propagation can be adapted directly in order to control the meaning of an explanation.
            An example of a local explanation using two kinds of \glspl{roi} are shown in Figure~\ref{fig:local_to_glocal_seg} (\emph{right}),
            where both the whole segmentation output of a bus or the bus wheels are explained via binary masking.

            For the initialization of relevance propagation at the output,
            we adhere to the in-literature often practiced approach by focusing on output scores corresponding to the highest class prediction.
            In order to weight relevance scores uniformly,
            the relevance propagation for explaining class $y$ is initialized in the last output layer $L$ as
            
            \begin{equation}
                R^L_{(p, q, c)}(\x|\theta) = \delta_{cy} \boldsymbol{1}_{(p,q)}(\x|y)
            \end{equation}
            with an indicator function
            \begin{equation}
                \boldsymbol{1}_{(p,q)}(\x|y) = 
                \begin{cases}
                      1, & \text{if}\ y=\argmax_{k}{f_{(p,q,k)}(\x)} \\
                      0, & \text{otherwise}~.
                \end{cases}
            \end{equation} 
             The tensor indices $p$ and $q$ refer here to the spatial dimension $w \times h$ and $c$ describing the channel dimension, \ie class confidence.
            Thus, not whole channel output maps are explained,
            but only the output values,
            which correspond to the selected class.
            
            \paragraph{Object Detection}
            
            Object detection networks often consist of a decoder part and a prediction module. 
            The output per bounding box then includes class scores and, in addition to segmentation models, information about the position of the bounding box.
            Local attributions are then similarly computed to the classification task \wrt the class score of a chosen bounding box.
            
            Regarding a bounding box predictor
            $f: \mathbb{R}^n \rightarrow \mathbb{R}^{N\times (n_c+4)}$ with an output of $N$ bounding boxes for $n_c$ object classes with four coordinates,
            the relevance $R_i^{(k)}(\x|\theta)$ of feature $x_i$ for bounding box $k$ of class $y$ is then given by initializing the relevance propagation at the output as
            
            \begin{equation}
                R^L_{(b, c)}(\x|\theta) = \delta_{bk} \delta_{cy} f_c(\x)
            \end{equation}
            with $b$ representing the bounding box axis.
            Thus,
            for each bounding box,
            an individual explanation can be computed, as is shown in Figure~\ref{fig:local_to_glocal_seg} (\emph{left}).

\section{Experiments}
\label{sec:experiments}

    The experimental section is divided into two parts, beginning with the evaluation of our concept-based \gls{xai} method using \gls{crp} and different feature attribution methods,
    thereby showing the superiority of computing latent relevance scores instead of activations -- especially in the multi-object case.
    In the second part,
    we leverage the fact that we can localize the object through ground truth masks or bounding boxes
    and thus are able to measure how much the background instead of the actual object is used.
    We show how this can be used to detect possible biases in the model corresponding to specific feature encodings.

    Concretely,
    we apply our method to PyTorch implementations of a UNet on the CityScapes dataset, a DeepLabV3+ on the Pascal VOC 2012 dataset,
    a YOLOv5 and YOLOv6 model on the MS COCO 2017 dataset.
    Please refer to Appendix~\ref{app:technical_details} for more details on the models.
    
    \subsection{Evaluation of Concept-based Explanations}
    \label{sec:explanation_quality}
    In recent years,
    several methods have been proposed to evaluate local explanations,
    that are readily applicable through unifying frameworks. 
    Following the authors of \cite{hedstrom2022quantus, chalasani2020concise}, we evaluate our presented method \wrt to faithfulness and complexity.
    Faithfulness measures whether an attribution truly represents features utilized by the model during inference, while complexity measures how concise explanations are, which is of interest in context of \eg human interpretation.
    Since the literature for evaluating local \emph{concept-based} explanations is limited, 
    we propose two experiments to test for faithfulness and complexity.

        \begin{table*}[h!]\centering
         \caption{Experimental results for evaluating various concept attribution approaches in terms of faithfulness (higher is better ($\uparrow$)) and complexity (lower is better ($\downarrow$)).
        For each approach, the evaluated scores for all models are displayed (UNet/DeepLabV3+/YOLOv5/YOLOv6). }
       \resizebox{1\textwidth}{!}{
            \begin{tabular*}{\textwidth}{@{}lcccc@{}}
            \toprule
                & \multicolumn{2}{c}{\textbf{faithfulness} ($\uparrow$)} & \multicolumn{2}{c}{\textbf{complexity} ($\downarrow$)} \\ \midrule
                & concept flipping   &  concept insertion   & explanation variation   & concepts for 80\,\% of attr. (\%)    \\ \midrule
              LRP-z$^+$ &       $4.57/2.52/1.85/1.20$           & $4.96/2.62/2.46/1.65$         & $0.47/0.19/0.27/0.39$           & $27.2/42.9/53.0/21.9$ \\
              LRP-$\gamma$ &    $\mathbf{4.90}/2.78/2.02/1.34$  & $5.54/2.75/2.60/\mathbf{1.72}$ & $0.66/0.28/0.47/0.56$          & $\mathbf{22.2}/\mathbf{34.1}/41.5/\mathbf{17.8}$ \\
              LRP-$\varepsilon$&$4.32/3.24/\mathbf{2.39}/\mathbf{1.45}$  & $5.18/3.27/\mathbf{2.78}/1.37$ & $1.12/0.65/0.73/0.94$ & $26.6/37.9/\mathbf{35.1}/22.5$ \\
              gradient    &     $4.81/\mathbf{3.66}/1.81/1.09$  & $\mathbf{5.65}/\mathbf{3.77}/2.38/1.19$ & $0.72/0.56/0.86/0.98$ & $32.5/42.6/37.0/27.3$ \\
              max-act.&   $3.03/2.10/1.46/0.83$           & $3.72/2.36/2.15/1.15$         & $\mathbf{0.28}/\mathbf{0.09}/\mathbf{0.15}/\mathbf{0.22}$ & $60.5/63.2/68.3/44.4$ \\ 
              mean-act.&  $2.91/2.11/1.43/1.00$           & $3.63/2.36/2.09/1.35$         & $0.77/0.21/0.21/0.29$           & $58.7/52.1/54.3/49.2$ \\ \bottomrule
            \end{tabular*}
        }
        \label{tab:experiments:faithfulness}
        \end{table*}

    \subsubsection{Faithfulness of Concept Relevances}
    \label{sec:experiments:faithfulness}
    In order to assess the faithfulness of our concept-based explanations, 
    we measure the impact on the decision outcome if a set of concepts is perturbed.
    This idea is analogous to the pixel flipping experiment in \cite{bach2015pixel},
    only to use latent concepts instead of input features.
    
    Concretely, 
    we begin by computing the relevance scores of all concepts in a layer for a given object prediction. 
    Please note,
    that since we conceptualize each convolutional channel to correspond to a distinct concept,
    the concept relevance of a channel is acquired via spatially sum-aggregation of intermediate relevance scores.
    Then, we successively deactivate the most relevant channels first in descending order by setting their activations to zero, re-evaluate the model output and measure output differences. 
    Whereas for object detection models the change in the class logit of a predicted bounding box is computed,
    for segmentation models the mean change of the output mask's logits is calculated.

    Alternatively, we also perform concept flipping backwards, by initializing all filters with zero activation
    and successively ``unflipping'' the most relevant concepts in descending order, \ie, restoring their original activation values.
    This technique is designated as ``concept insertion''.

    \begin{figure} 
        \centering
            \includegraphics[width=1\linewidth]{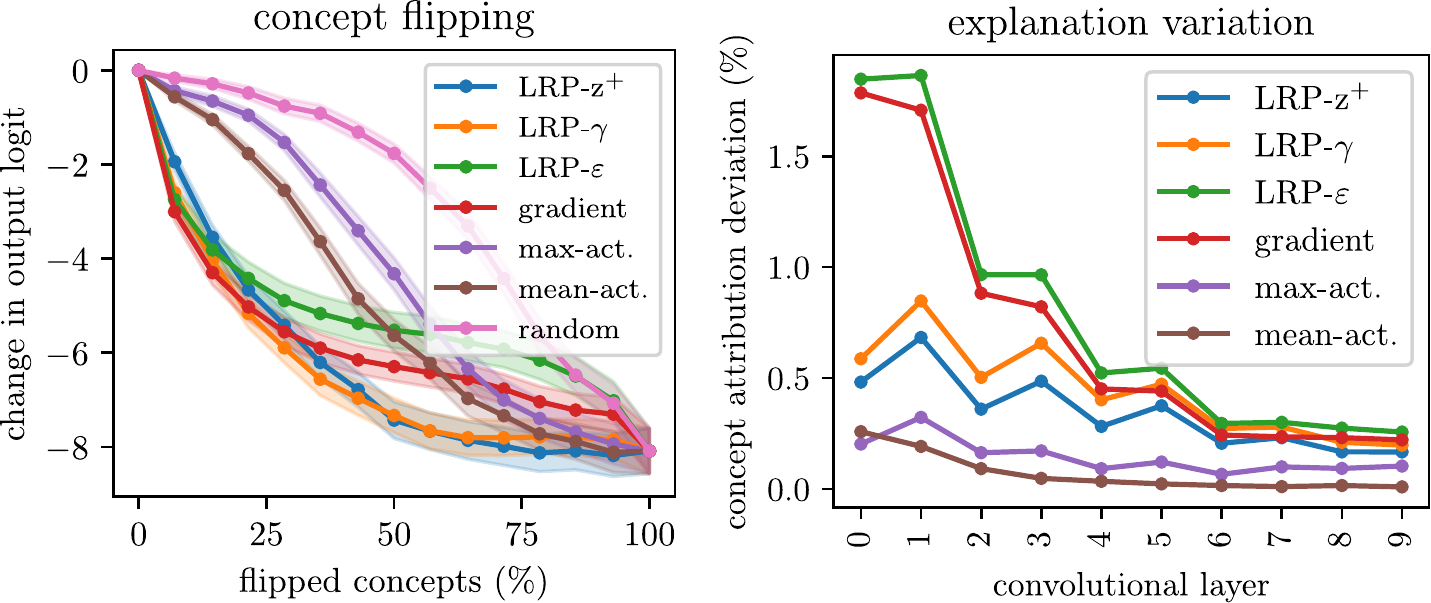}
            \captionof{figure}{Measuring the faithfulness (\emph{left}) and complexity (\emph{right}) of concept attributions for the UNet architecture.
            (\emph{Left}): Activations of the most relevant (or activating) concepts are set to zero successively, and the output difference measured. The \gls{sem} over 100 samples is shown in semi-transparent color.
            (\emph{Right}): The variation of concept attributions is lowest using activations and highest using gradient or \gls{lrp}-$\varepsilon$ relevances. 
            Shown are the first ten convolutional layers of the UNet model.
            }
            \label{fig:experiments:explanation_quality:complexity_faithfulness_example}
    \end{figure}

        A faithful explanation is hereby characterized by a strong decline in performance if concepts are flipped,
        or a strong incline in performance if concepts are inserted.
        A typical experimental outcome is shown in Figure~\ref{fig:experiments:explanation_quality:complexity_faithfulness_example} (\emph{left}).
        Here different \gls{lrp}-rules, pure gradient, activation (mean and sum-pooling), and a random perturbation baseline are compared
        for concepts in layer \texttt{features.10} of the UNet model.
        As demonstrated,
        perturbation of the most activating concepts is significantly less faithful than using relevance-based scores.
        This is expected,
        since relevance is object-specific and thereby filters out other activating concepts irrelevant for the current detection.
        Regarding concept relevances,
        the scores of gradient and \gls{lrp}-$\varepsilon$ are often characterized by the strongest decline/incline,
        as the gradient of the model faithfully measures the local sensitivity of the model to changes.
        However,
        as more features are perturbed,
        \gls{lrp}-$\gamma$ and \gls{lrp}-$z^+$ often perform better,
        as they better represent the important features in a more global manner by filtering out noisy attributions \cite{bach2015pixel}.
        In order to receive a score for a whole model,
        the faithfulness tests are performed in various layers throughout the models on 100 randomly chosen predictions,
        and the area under or over the curve measured per layer and mean-aggregated to form a final faithfulness score.
        As can be seen in Table~\ref{tab:experiments:faithfulness},
        the results depicted in Figure~\ref{fig:experiments:explanation_quality:complexity_faithfulness_example} are reflected throughout all tested models, by mostly showing the best scores for gradient and \gls{lrp}-$\varepsilon$.
        Layer-wise faithfulness scores for all models can be found in the
        Appendix~\ref{app:faithfulness}.
        
    \subsubsection{Explanation Complexity and Interpretation Workload}
    \label{sec:experiments:complexity}
    While high faithfulness suggests that the concept attributions represent the model behavior correctly, 
    they can still be noisy and not human-interpretable \cite{kohlbrenner2020towards}. 
    This effect is due to highly non-linear decision boundaries in \glspl{dnn} \cite{balduzzi2017shattered}.
    In order to measure the complexity of explanations and the workload a stakeholder has to put in for understanding the explanations, two different measures are computed.

    First,
    the standard deviation of latent concept attributions per class is measured, indicating the amount of noise.
    A low variation suggests,
    that explanations of the same class are similar and precise, resulting in a lower amount of complexity.
    As a second measure,
    the amount of concepts necessary to study in order to comprehend 80\,\% of all attributions is computed.
    The more relevance is focused on a small number of concepts, the fewer concepts need to be analyzed.

    An example for measuring explanation complexity is shown in Figure~\ref{fig:experiments:explanation_quality:complexity_faithfulness_example} (\emph{right}),
    where the explanation variation in the first ten convolutional layers of the UNet architecture is shown.
    Activations exhibit the smallest variation because filters activate on average more often if the feature is present in the image even though not used for inference.
    Regarding relevance-based concept attributions,
    gradient and \gls{lrp}-$\varepsilon$ show especially in lower-level layers a high deviation,
    indicating noisy attributions.

    Whereas activation-based approaches lead to a low explanation variation,
    the distribution of concept attributions is rather uniform,
    leading to unconcise explanations and a large interpretation workload.
    Here,
    relevance-based approaches,
    which generate object-specific attributions result in smaller relevant concept sets.
    Plots for all architectures are illustrated in Appendix~\ref{app:complexity}.
    
    The results of measuring explanation complexity for every model using all predictions in the test datasets
    are given in Table~\ref{tab:experiments:faithfulness},
    confirming the previous observations in Figure~\ref{fig:experiments:explanation_quality:complexity_faithfulness_example}.
    
    Taking into account the results of the faithfulness tests,
    it is apparent,
    that relevance-based concept attributions show higher faithfulness than activation,
    but are not necessarily easier to interpret in terms of explanation complexity alone.
    Here,
    \gls{lrp}-$\gamma$ attributions show a good compromise between faithfulness and complexity in most experiments.

    \subsection{Concept Context Scores for Bias Detection}
    \label{sec:context}
        In an ideal world,
        a \gls{dnn} trained with enough variety in training data learns abstract and generalized features.
        However,
        this is not always the case and several works have shown that \glspl{dnn} develop biases or Clever Hans features,
        \eg detecting specific classes because of correlations in the data, such as watermark signs \cite{stock2018convnets, schramowski2020making, anders2022finding}.

        \begin{figure} 
            \centering
                 \includegraphics[width=0.9\linewidth]{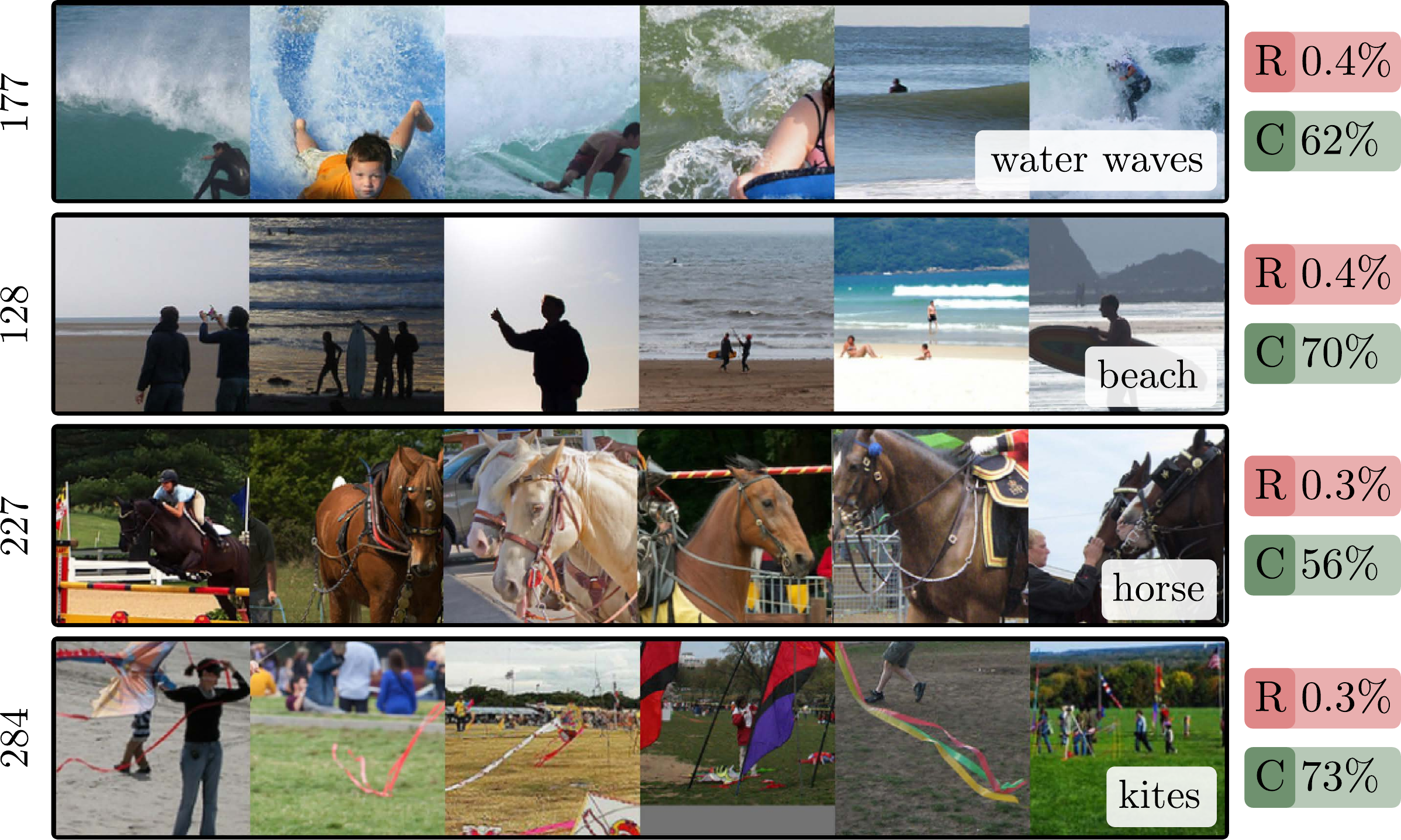}
                \captionof{figure}{Examples of background concepts for class ``person''  of the YOLOv6 model in layer \texttt{ERBlock\_5.0.rbr\_dense} with 512 concepts. For each concept, the mean relevance (R), context score (C) and context interpretation (white box) is given.}
                \label{fig:experiments:context:person_background_concepts_example}
        \end{figure}
        
        Also in our experimental datasets,
        objects are often displayed in the same environments or together with other object classes.
        A person is often pictured together with a kite, sitting on a horse, walking on the beach, or standing on a surfing board in splashing water.
        In fact,
        concepts can be found in the YOLOv6 model for exactly the above-mentioned use cases, 
        as shown in Figure~\ref{fig:experiments:context:person_background_concepts_example}.

        \subsubsection{Measuring the Context of Concepts}
        In order to reveal to what degree the model uses concepts for background features,
        we in the following compare computation of context scores by three different means:
        (1) latent concept activation map,
        (2) latent concept relevance map and
        (3) concept-conditional \gls{lcrp} heatmap.
        We define the context score $C$ of concept $i$ as the fraction of positive attribution $a^+$ outside of the object bounding box compared to the overall sum, as 
        \begin{equation}
        \label{eq:experiments:context:context_score}
         C_i = \sum_j^n \frac{1}{n} \frac{\sum_{p, q} a^+_{(p,q,i)}(\x_j) \bar m_{(p,q)}(\x_j)}{a_i^+(\x_j)} \in [0, 1]
        \end{equation}
        for $n$ samples $\x_j$ and mask $\bar m\in\{0,1\}$ marking all background values with a value of one and zero else.
        Indices $p$ and $q$ refer to the spatial dimension of the attribution maps.
        In case of segmentation,
        the object is localized through pixel-accurate ground-truth masks,
        whereas for object detection bounding boxes are available.
        For measuring the context score in the latent space,
        masks or boxes are resized.

        Please note,
        that the use of latent feature maps is only possible for convolutional layers,
        but not for dense layers, contained in most classification networks.
        \gls{lcrp} heatmaps, however, can be computed for both layer types.

        \begin{figure}  
            \centering
                \includegraphics[width=0.9\linewidth]{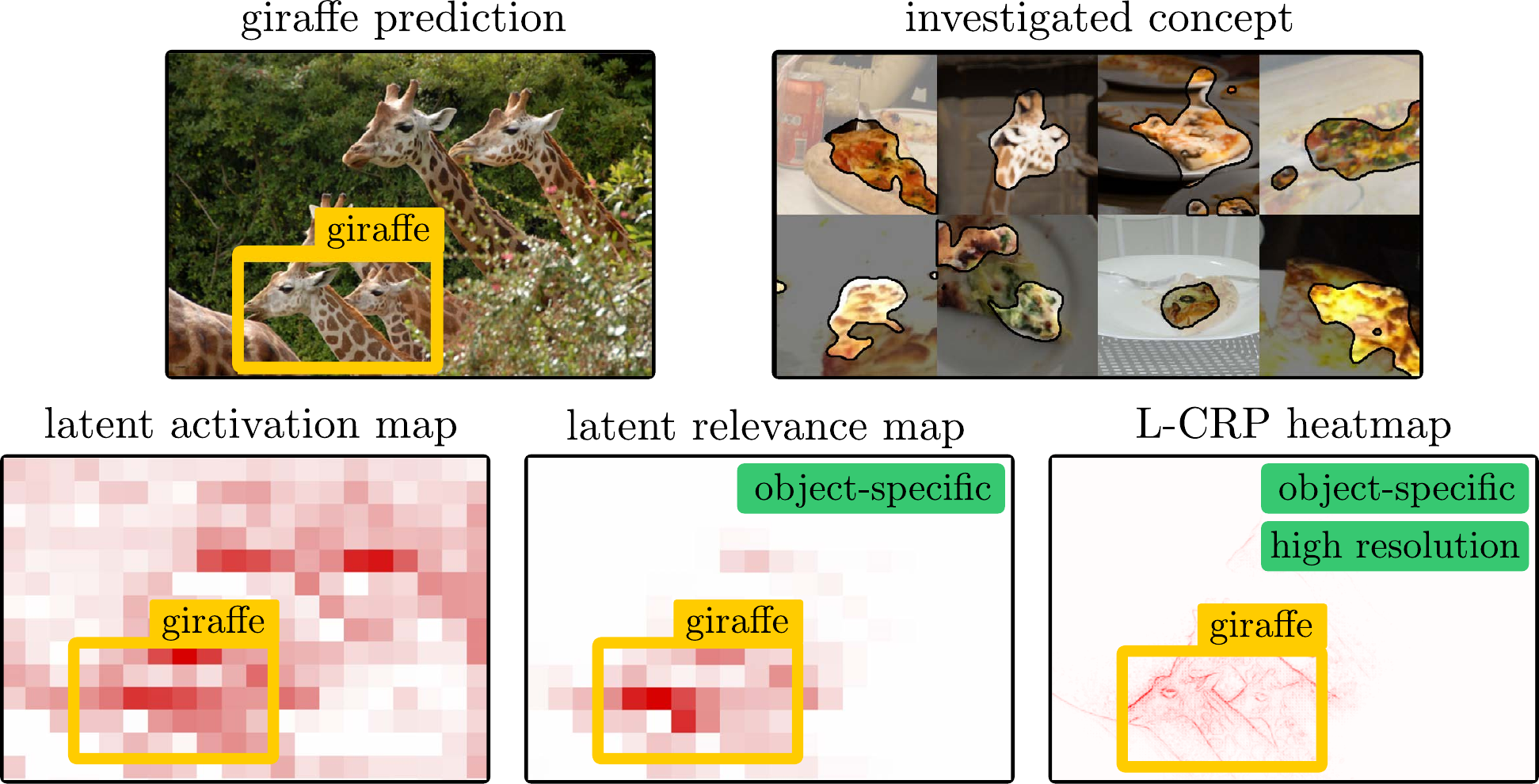}
                \captionof{figure}{Measuring the context score of a concept encoding for the spotted texture of the giraffe skin (shown are most relevant concept samples) using the latent relevance and activation map as well as \gls{lcrp} heatmap. 
                The concept activates broadly on all giraffes in the image, thus leading to an over-estimation of context using latent activations.
                \gls{lcrp} results in the most precise localization and is object-specific at the same time.
                }
                \label{fig:experiments:context:measuring}
        \end{figure}
        An example of computing the context score is shown in Figure~\ref{fig:experiments:context:measuring} for the YOLOv6 model and concept 28 of layer \texttt{ERBlock\_5.0.rbr\_dense},
        which corresponds to the skin pattern of the giraffe.
        As can be seen,
        latent activation maps are not optimal for measuring context scores,
        as they are not object- and outcome-specific, leading to a broad activation on all giraffes in the image.
        Thus,
        context scores are over-estimated using activation.
        It is further apparent, 
        that for high-level layers the high resolution of \gls{lcrp} maps leads to a more accurate context estimation than low-resolution latent maps.

    \begin{figure} 
        \centering
            \includegraphics[width=1\linewidth,trim={0cm 0 0cm 0.6cm},clip]{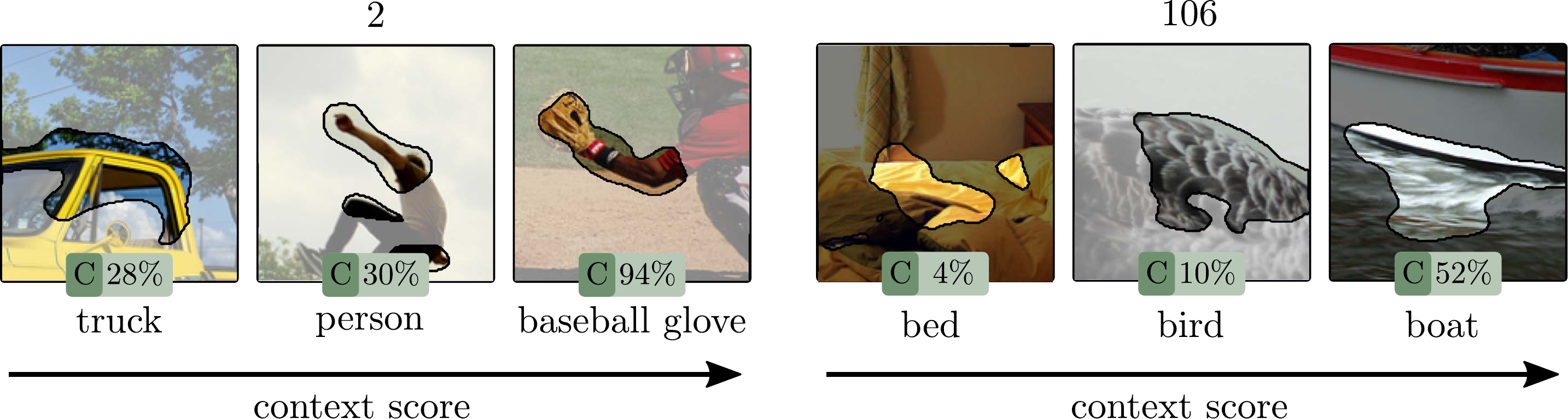}
            \captionof{figure}{Context scores (C) of concepts vary between classes, as shown for concepts 2 (\emph{left}) and 106 (\emph{right}) of the YOLOv6 model in layer \texttt{ERBlock\_5.0.rbr\_dense}. }
            \label{fig:experiments:context:context_varys_classes}
    \end{figure}
        When computing context scores,
        we found that concepts are used differently throughout object classes, as shown in Figure~\ref{fig:experiments:context:context_varys_classes}.
        Here,
        \eg, the texture of waves can be found in blankets of beds, feathers for birds, or in the water surrounding boats.
        This illustrates,
        that it is crucial to measure the context of concepts for each class.

    \subsubsection{Evaluating Context Scores}

        For evaluating estimated context scores,
        we propose to measure the model's sensitivity of concepts on the background of objects.
        We therefore inspect the influence on the concepts' relevances
        when the background is perturbed.
        Concretely,
        the background sensitivity $S$ of concept $i$ is defined as
        \begin{equation}
        \label{eq:experiments:context:background_sensitivity}
            S_i = \sum_j^n \frac{1}{n} \frac{|R_i(\x_j) - R_i(\Tilde{\x}_j)|}{\max \{|R_i(\x_j)|, |R_i(\Tilde{\x}_j)|\}} \in [0, 1]\,,
        \end{equation}
        with concept relevance $R_i(\x_j)$ and object samples $\Tilde{\x}_j$ and $\x_j$ with and without perturbed backgrounds, respectively.
        Specifically, we applied gray-scale random noise, random noise, gray, and random color perturbation.
        Each perturbation is further performed with 100\,\% and 50\,\% alpha-blending, totaling 8 perturbations on 60 random detections.
        
        The context scores are computed for the 50 most relevant concepts of a class and the corresponding 15 most relevant detections, according to \gls{lcrp} attributions with the \gls{lrp}-z$^+$-rule on the test data.

        \begin{table}[]
        \centering
            \caption{Comparing computed context scores with measured background sensitivity. Ideal is a high correlation and low Root Mean Square Deviation (RMSD). Correlation values are given for all models (UNet/DeepLabV3+/YOLOv5/YOLOv6).}
            \begin{tabular}{@{}lcccc@{}}
            \toprule
                          & correlation & RMSD \\ \midrule
            L-\gls{crp} (ours)   &   $\mathbf{0.55}/\mathbf{0.68}/\mathbf{0.80}/\mathbf{0.71}$  & $\mathbf{0.17}$ \\
            latent relevance  &   $0.53/0.67/0.71/0.70$   &     $0.20$         \\ 
            latent activation &    $0.50/0.63/0.71/0.66$  &      $0.40$          \\\bottomrule
            \end{tabular}
        \label{tab:experiments:context:evaluation}
        \end{table}
        
        Ideally,
        concepts with a high context score $C$ also have a high background sensitivity $S$.
        As is summarized in Table~\ref{tab:experiments:context:evaluation},
        \gls{lcrp} results in both the highest correlation and lowest Root Mean Square Deviation values between context and sensitivity scores. 
        Here, concept scores are evaluated for three layers of each model.
        It is to note,
        that using activations leads to relatively high correlations, but large RMSDs as context scores are over-estimated.
        This is further illustrated by distribution plots in the Appendix~\ref{app:context}.

    \subsubsection{Context-based Interaction with the Model}

        The availability of context scores in combination with our method allows to detect the use of single background features,
        and thus to precisely interact with the model.

        \begin{figure} 
        \centering
            \includegraphics[width=1\linewidth]{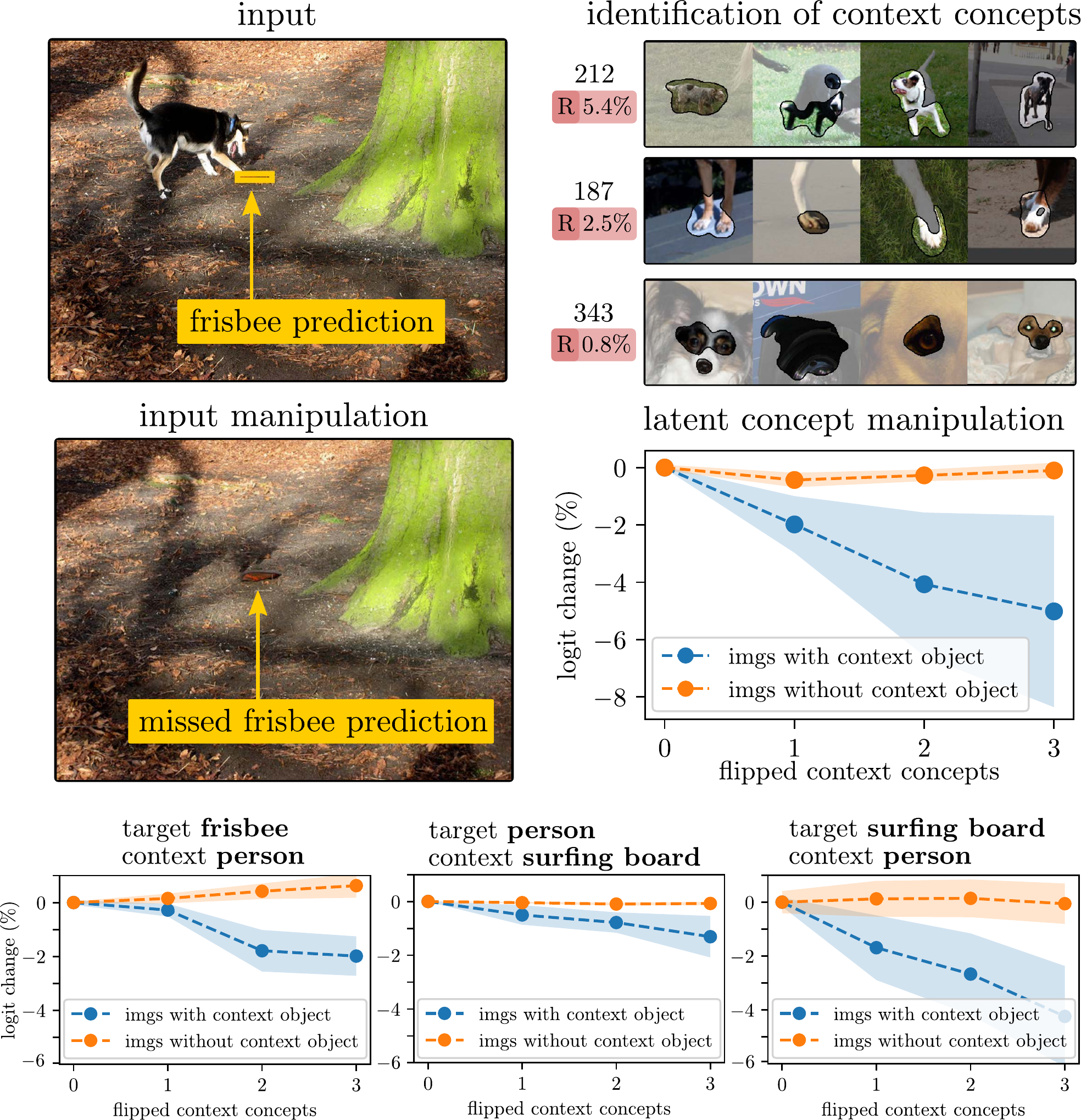}
            \captionof{figure}{Dog concepts are relevant for a frisbee detection (\emph{top}),
            leading to a missed prediction by the YOLOv6 model when the dog is removed from the image.
            We alternatively flip the dog concepts and measure a significant decrease in the output of the frisbee class for samples with both frisbee and dog present. We performe latent concept flipping also for other cases (\emph{bottom}). The \gls{sem} is visualized in semi-transparent color.
            Concept visualizations of all flipped background concepts can be found in the Appendix~\ref{app:context}.}
            \label{fig:dog_frisbee_perturbation}
    \end{figure}
        
        To probe the reaction of the model, the object's background can be manipulated.
        As an example,
        we noticed that frisbee detections rely on background concepts corresponding to features of dogs, shown in Figure~\ref{fig:dog_frisbee_perturbation}.
        Removing the dog from the image via in-painting leads then to a missed detection of the YOLOv6 model.
        
        Alternatively,
        as we are able to pinpoint background concepts in the latent space,
        we can also flip the corresponding concepts,
        and measure the effect on the prediction.
        When a dog is present in images,
        concept flipping decreases the predicted output logit by about 6\,\%, as shown in Figure~\ref{fig:dog_frisbee_perturbation},
        whereas predictions without any dog present are not influenced significantly.
        
        We further perform latent background concept flipping for frisbee or surfing board detections with ``person'' context features
        as well as person detections with ``surfing board'' concepts,
        visualized in Figure~\ref{fig:dog_frisbee_perturbation} (\emph{bottom}).
        Here,
        the removal of person concepts affects the surfing board prediction more strongly than the other way around.
        We can even find examples where flipping three background concepts leads to missed surfing board predictions (shown in Appendix~\ref{app:context}).
        This is expected,
        as surfing boards are more likely to be depicted with a person (97\,\% co-occurrence) than a person with a surfing board (5\,\% co-occurrence) in the training data, 
        favoring the likeliness of the model to use context features.
        The variety of ``person'' contexts is much higher,
        making it possible for the model to become more generalized in concept utility \cite{lapuschkin2019unmasking}.
        The visualization of all flipped background concepts can be found in the Appendix~\ref{app:context}.

\section{Conclusion}
We propose \gls{lcrp} as an extension of the recent \gls{crp} method, enabling local concept-based understanding of segmentation and object detection models. 
By visualizing and localizing the model's internal representations utilizing \gls{lcrp}, 
the insight that a stakeholder gains can be significantly improved compared to traditional local XAI methods,
which tend to merely resemble the object localization output.
We evaluate several latent concept attribution methods showing a trade-off between explanation faithfulness and complexity,
with relevance-based attributions providing the best compromise.
Finally, we apply our method to measure the amount of context by which concepts are used by a model for the prediction of different object classes.
We validate our results and show how context scores can be leveraged to reveal context bias and interact with the model.

\section*{Acknowledgements}
This work was partly supported by the German Ministry for Education and Research (BMBF) under Grants [BIFOLD (01IS18025A, 01IS18037I) and Patho234 (031L0207C)], the European Union's Horizon 2020 research and innovation programme as grant [iToBoS (965221)],  and the state of Berlin within the innovation support program ProFIT as grant [BerDiBa (10174498)].

{\small
\bibliographystyle{ieee_fullname}
\bibliography{main}

\begin{thebibliography}{10}\itemsep=-1pt

\bibitem{achtibat2022towards}
Reduan Achtibat, Maximilian Dreyer, Ilona Eisenbraun, Sebastian Bosse, Thomas
  Wiegand, Wojciech Samek, and Sebastian Lapuschkin.
\newblock From ``where" to ``what": Towards human-understandable explanations
  through concept relevance propagation.
\newblock {\em arXiv preprint arXiv:2206.03208}, 2022.

\bibitem{ahmed2021explainable}
Awadelrahman~MA Ahmed and Leen~AM Ali.
\newblock Explainable medical image segmentation via generative adversarial
  networks and layer-wise relevance propagation.
\newblock {\em Nordic Machine Intelligence}, 1(1):20--22, 2021.

\bibitem{andeol2021learning}
L{\'e}o And{\'e}ol, Yusei Kawakami, Yuichiro Wada, Takafumi Kanamori,
  Klaus-Robert M{\"u}ller, and Gr{\'e}goire Montavon.
\newblock Learning domain invariant representations by joint wasserstein
  distance minimization.
\newblock {\em arXiv preprint arXiv:2106.04923}, 2021. 

\bibitem{anders2021software}
Christopher~J. Anders, David Neumann, Wojciech Samek, Klaus-Robert
  M{\"{u}}ller, and Sebastian Lapuschkin.
\newblock Software for dataset-wide {XAI:} from local explanations to global
  insights with {Zennit, CoRelAy, and ViRelAy}.
\newblock {\em arXiv preprint arXiv:2106.13200}, 2021.

\bibitem{anders2022finding}
Christopher~J Anders, Leander Weber, David Neumann, Wojciech Samek,
  Klaus-Robert M{\"u}ller, and Sebastian Lapuschkin.
\newblock Finding and removing clever hans: Using explanation methods to debug
  and improve deep models.
\newblock {\em Information Fusion}, 77:261--295, 2022.

\bibitem{arnold2019survey}
Eduardo Arnold, Omar~Y Al-Jarrah, Mehrdad Dianati, Saber Fallah, David Oxtoby,
  and Alex Mouzakitis.
\newblock A survey on 3d object detection methods for autonomous driving
  applications.
\newblock {\em IEEE Transactions on Intelligent Transportation Systems},
  20(10):3782--3795, 2019.

\bibitem{bach2015pixel}
Sebastian Bach, Alexander Binder, Gr{\'e}goire Montavon, Frederick Klauschen,
  Klaus-Robert M{\"u}ller, and Wojciech Samek.
\newblock On pixel-wise explanations for non-linear classifier decisions by
  {Layer-Wise Relevance Propagation}.
\newblock {\em PLoS ONE}, 10(7):e0130140, 2015.

\bibitem{balduzzi2017shattered}
David Balduzzi, Marcus Frean, Lennox Leary, J.~P. Lewis, Kurt~Wan{-}Duo Ma, and
  Brian McWilliams.
\newblock The shattered gradients problem: If resnets are the answer, then what
  is the question?
\newblock In {\em 34th International Conference on Machine Learning (ICML)},
  volume~70, pages 342--350, 2017.

\bibitem{bau2020understanding}
David Bau, Jun{-}Yan Zhu, Hendrik Strobelt, {\`{A}}gata Lapedriza, Bolei Zhou,
  and Antonio Torralba.
\newblock Understanding the role of individual units in a deep neural network.
\newblock {\em Proc. Natl. Acad. Sci. {USA}}, 117(48):30071--30078, 2020.

\bibitem{binder2020notes}
Alexander Binder.
\newblock Notes on canonization for resnets and densenets.
\newblock {https://github.com/AlexBinder/LRP\_Pytorch\_Resnets\_
  Densenet/blob/master/canonization\_doc.pdf}, 2020.

\bibitem{cammarata2020thread}
Nick Cammarata, Shan Carter, Gabriel Goh, Chris Olah, Michael Petrov, Ludwig
  Schubert, Chelsea Voss, Ben Egan, and Swee~Kiat Lim.
\newblock Thread: Circuits.
\newblock {\em Distill}, 5(3):e24, 2020.

\bibitem{chalasani2020concise}
Prasad Chalasani, Jiefeng Chen, Amrita~Roy Chowdhury, Xi Wu, and Somesh Jha.
\newblock Concise explanations of neural networks using adversarial training.
\newblock In {\em International Conference on Machine Learning}, pages
  1383--1391. PMLR, 2020.

\bibitem{chen2018encoder}
Liang-Chieh Chen, Yukun Zhu, George Papandreou, Florian Schroff, and Hartwig
  Adam.
\newblock Encoder-decoder with atrous separable convolution for semantic image
  segmentation.
\newblock In {\em Proceedings of the European conference on computer vision
  (ECCV)}, pages 801--818, 2018.

\bibitem{eu2019building}
{Commission to the European Parliament}, {the Council}, {the European Economic
  and Social Committee}, and {the Committee of the Regions}.
\newblock Communication: Building trust in human centric artificial
  intelligence.
\newblock {\em COM}, 168, 2019.

\bibitem{cordts2016cityscapes}
Marius Cordts, Mohamed Omran, Sebastian Ramos, Timo Rehfeld, Markus Enzweiler,
  Rodrigo Benenson, Uwe Franke, Stefan Roth, and Bernt Schiele.
\newblock The cityscapes dataset for semantic urban scene understanding.
\newblock In {\em Proceedings of the IEEE conference on computer vision and
  pattern recognition}, pages 3213--3223, 2016.

\bibitem{ding2021repvgg}
Xiaohan Ding, Xiangyu Zhang, Ningning Ma, Jungong Han, Guiguang Ding, and Jian
  Sun.
\newblock Repvgg: Making vgg-style convnets great again.
\newblock In {\em Proceedings of the IEEE/CVF Conference on Computer Vision and
  Pattern Recognition}, pages 13733--13742, 2021.

\bibitem{jocher2021yolov5}
Glenn~Jocher et. al.
\newblock {YOLOv5n Nano models, Roboflow integration, TensorFlow export, OpenCV
  DNN support}.
\newblock {\em Zenodo}, 2021.

\bibitem{evans2022explainability}
Theodore Evans, Carl~Orge Retzlaff, Christian Gei{\ss}ler, Michaela Kargl,
  Markus Plass, Heimo M{\"u}ller, Tim-Rasmus Kiehl, Norman Zerbe, and Andreas
  Holzinger.
\newblock The explainability paradox: Challenges for xai in digital pathology.
\newblock {\em Future Generation Computer Systems}, 133:281--296, 2022.

\bibitem{everingham2010pascal}
Mark Everingham, Luc Van~Gool, Christopher~KI Williams, John Winn, and Andrew
  Zisserman.
\newblock The pascal visual object classes (voc) challenge.
\newblock {\em International journal of computer vision}, 88(2):303--338, 2010.

\bibitem{falk2019u}
Thorsten Falk, Dominic Mai, Robert Bensch, {\"O}zg{\"u}n {\c{C}}i{\c{c}}ek,
  Ahmed Abdulkadir, Yassine Marrakchi, Anton B{\"o}hm, Jan Deubner, Zoe
  J{\"a}ckel, Katharina Seiwald, et~al.
\newblock U-net: deep learning for cell counting, detection, and morphometry.
\newblock {\em Nature methods}, 16(1):67--70, 2019.

\bibitem{fang2019deeplabv3plus}
Gongfan Fang.
\newblock Deeplabv3plus-pytorch.
\newblock \url{https://github.com/VainF/DeepLabV3Plus-Pytorch}, 2019.

\bibitem{goh2021multimodal}
Gabriel Goh, Nick Cammarata, Chelsea Voss, Shan Carter, Michael Petrov, Ludwig
  Schubert, Alec Radford, and Chris Olah.
\newblock Multimodal neurons in artificial neural networks.
\newblock {\em Distill}, 6(3):e30, 2021.

\bibitem{goodman2017european}
Bryce Goodman and Seth Flaxman.
\newblock European union regulations on algorithmic decision-making and a
  “right to explanation”.
\newblock {\em AI magazine}, 38(3):50--57, 2017.

\bibitem{gudovskiy2018explain}
Denis Gudovskiy, Alec Hodgkinson, Takuya Yamaguchi, Yasunori Ishii, and Sotaro
  Tsukizawa.
\newblock Explain to fix: A framework to interpret and correct dnn object
  detector predictions.
\newblock {\em arXiv preprint arXiv:1811.08011}, 2018.

\bibitem{guillemot2020breaking}
Mathilde Guillemot, Catherine Heusele, Rodolphe Korichi, Sylvianne Schnebert,
  and Liming Chen.
\newblock Breaking batch normalization for better explainability of deep neural
  networks through {Layer-Wise Relevance Propagation}.
\newblock {\em arXiv preprint arXiv:2002.11018}, 2020.

\bibitem{he2016deep}
Kaiming He, Xiangyu Zhang, Shaoqing Ren, and Jian Sun.
\newblock Deep residual learning for image recognition.
\newblock In {\em Proceedings of the IEEE conference on computer vision and
  pattern recognition}, pages 770--778, 2016.

\bibitem{hedstrom2022quantus}
Anna Hedstr{\"o}m, Leander Weber, Dilyara Bareeva, Franz Motzkus, Wojciech
  Samek, Sebastian Lapuschkin, and Marina M-C H{\"o}hne.
\newblock Quantus: an explainable ai toolkit for responsible evaluation of
  neural network explanations.
\newblock {\em arXiv preprint arXiv:2202.06861}, 2022.

\bibitem{hoyer2019grid}
Lukas Hoyer, Mauricio Munoz, Prateek Katiyar, Anna Khoreva, and Volker Fischer.
\newblock Grid saliency for context explanations of semantic segmentation.
\newblock {\em Advances in neural information processing systems}, 32, 2019.

\bibitem{hui2019batchnorm}
Lucas Y.~W. Hui and Alexander Binder.
\newblock {{BatchNorm Decomposition}} for deep neural network interpretation.
\newblock In {\em Advances in Computational Intelligence}, Lecture Notes in
  Computer Science, pages 280--291, Cham, 2019. {Springer}.

\bibitem{Iakubovskii:2019}
Pavel Iakubovskii.
\newblock Segmentation models pytorch.
\newblock \url{https://github.com/qubvel/segmentation_models.pytorch}, 2019.

\bibitem{karasmanoglou2022heatmap}
Apostolos Karasmanoglou, Marios Antonakakis, and Michalis Zervakis.
\newblock Heatmap-based explanation of yolov5 object detection with layer-wise
  relevance propagation.
\newblock In {\em 2022 IEEE International Conference on Imaging Systems and
  Techniques (IST)}, pages 1--6. IEEE, 2022.

\bibitem{karri2022explainable}
Meghana Karri, Chandra Sekhara~Rao Annavarapu, and U~Rajendra Acharya.
\newblock Explainable multi-module semantic guided attention based network for
  medical image segmentation.
\newblock {\em Computers in Biology and Medicine}, page 106231, 2022.

\bibitem{kawauchi2022shap}
Hiroki Kawauchi and Takashi Fuse.
\newblock Shap-based interpretable object detection method for satellite
  imagery.
\newblock {\em Remote Sensing}, 14(9):1970, 2022.

\bibitem{kim2018interpretability}
Been Kim, Martin Wattenberg, Justin Gilmer, Carrie Cai, James Wexler, Fernanda
  Viegas, et~al.
\newblock Interpretability beyond feature attribution: Quantitative testing
  with concept activation vectors ({TCAV}).
\newblock In {\em 35th International Conference on Machine Learning (ICML)},
  pages 2668--2677, 2018.

\bibitem{kohlbrenner2020towards}
Maximilian Kohlbrenner, Alexander Bauer, Shinichi Nakajima, Alexander Binder,
  Wojciech Samek, and Sebastian Lapuschkin.
\newblock Towards best practice in explaining neural network decisions with
  {LRP}.
\newblock In {\em 2020 International Joint Conference on Neural Networks
  (IJCNN)}, pages 1--7. IEEE, 2020.

\bibitem{koker2021u}
Teddy Koker, Fatemehsadat Mireshghallah, Tom Titcombe, and Georgios Kaissis.
\newblock U-noise: Learnable noise masks for interpretable image segmentation.
\newblock In {\em 2021 IEEE International Conference on Image Processing
  (ICIP)}, pages 394--398. IEEE, 2021.

\bibitem{lapuschkin2019unmasking}
Sebastian Lapuschkin, Stephan W{\"a}ldchen, Alexander Binder, Gr{\'e}goire
  Montavon, Wojciech Samek, and Klaus-Robert M{\"u}ller.
\newblock Unmasking clever hans predictors and assessing what machines really
  learn.
\newblock {\em Nature Communications}, 10(1):1--8, 2019.

\bibitem{li2022yolov6}
Chuyi Li, Lulu Li, Hongliang Jiang, Kaiheng Weng, Yifei Geng, Liang Li, Zaidan
  Ke, Qingyuan Li, Meng Cheng, Weiqiang Nie, et~al.
\newblock Yolov6: a single-stage object detection framework for industrial
  applications.
\newblock {\em arXiv preprint arXiv:2209.02976}, 2022.

\bibitem{lin2014microsoft}
Tsung-Yi Lin, Michael Maire, Serge Belongie, James Hays, Pietro Perona, Deva
  Ramanan, Piotr Doll{\'a}r, and C~Lawrence Zitnick.
\newblock Microsoft coco: Common objects in context.
\newblock In {\em European conference on computer vision}, pages 740--755.
  Springer, 2014.

\bibitem{losch2021semantic}
Max Losch, Mario Fritz, and Bernt Schiele.
\newblock Semantic bottlenecks: Quantifying and improving inspectability of
  deep representations.
\newblock {\em International Journal of Computer Vision}, 129(11):3136--3153,
  2021.

\bibitem{mankodiya2022od}
Harsh Mankodiya, Dhairya Jadav, Rajesh Gupta, Sudeep Tanwar, Wei-Chiang Hong,
  and Ravi Sharma.
\newblock Od-xai: Explainable ai-based semantic object detection for autonomous
  vehicles.
\newblock {\em Applied Sciences}, 12(11):5310, 2022.

\bibitem{mintz2019introduction}
Yoav Mintz and Ronit Brodie.
\newblock Introduction to artificial intelligence in medicine.
\newblock {\em Minimally Invasive Therapy \& Allied Technologies},
  28(2):73--81, 2019.

\bibitem{montavon2019layer}
Gr{\'{e}}goire Montavon, Alexander Binder, Sebastian Lapuschkin, Wojciech
  Samek, and Klaus-Robert M{\"{u}}ller.
\newblock {Layer-Wise Relevance Propagation}: An overview.
\newblock In {\em Explainable {AI:} Interpreting, Explaining and Visualizing
  Deep Learning}, volume 11700 of {\em Lecture Notes in Computer Science},
  pages 193--209. Springer, Cham, 2019.

\bibitem{motzkus2022measurably}
Franz Motzkus, Leander Weber, and Sebastian Lapuschkin.
\newblock Measurably stronger explanation reliability via model canonization.
\newblock {\em arXiv preprint arXiv:2202.06621}, 2022.

\bibitem{olah2017feature}
Chris Olah, Alexander Mordvintsev, and Ludwig Schubert.
\newblock Feature visualization.
\newblock {\em Distill}, 2(11):e7, 2017.

\bibitem{olah2018building}
Chris Olah, Arvind Satyanarayan, Ian Johnson, Shan Carter, Ludwig Schubert,
  Katherine Ye, and Alexander Mordvintsev.
\newblock The building blocks of interpretability.
\newblock {\em Distill}, 3(3):e10, 2018.

\bibitem{petsiuk2021black}
Vitali Petsiuk, Rajiv Jain, Varun Manjunatha, Vlad~I Morariu, Ashutosh Mehra,
  Vicente Ordonez, and Kate Saenko.
\newblock Black-box explanation of object detectors via saliency maps.
\newblock In {\em Proceedings of the IEEE/CVF Conference on Computer Vision and
  Pattern Recognition}, pages 11443--11452, 2021.

\bibitem{radford2017learning}
Alec Radford, Rafal Jozefowicz, and Ilya Sutskever.
\newblock Learning to generate reviews and discovering sentiment.
\newblock {\em arXiv preprint arXiv:1704.01444}, 2017.

\bibitem{ronneberger2015u}
Olaf Ronneberger, Philipp Fischer, and Thomas Brox.
\newblock U-net: Convolutional networks for biomedical image segmentation.
\newblock In {\em International Conference on Medical image computing and
  computer-assisted intervention}, pages 234--241. Springer, 2015.

\bibitem{rudin2019stop}
Cynthia Rudin.
\newblock Stop explaining black box machine learning models for high stakes
  decisions and use interpretable models instead.
\newblock {\em Nature Machine Intelligence}, 1(5):206--215, 2019.

\bibitem{russakovsky2015imagenet}
Olga Russakovsky, Jia Deng, Hao Su, Jonathan Krause, Sanjeev Satheesh, Sean Ma,
  Zhiheng Huang, Andrej Karpathy, Aditya Khosla, Michael Bernstein, et~al.
\newblock Imagenet large scale visual recognition challenge.
\newblock {\em International journal of computer vision}, 115(3):211--252,
  2015.

\bibitem{samek2021explaining}
Wojciech Samek, Gr{\'e}goire Montavon, Sebastian Lapuschkin, Christopher~J
  Anders, and Klaus-Robert M{\"u}ller.
\newblock Explaining deep neural networks and beyond: A review of methods and
  applications.
\newblock {\em Proceedings of the IEEE}, 109(3):247--278, 2021.

\bibitem{schinagl2022occam}
David Schinagl, Georg Krispel, Horst Possegger, Peter~M Roth, and Horst
  Bischof.
\newblock Occam's laser: Occlusion-based attribution maps for 3d object
  detectors on lidar data.
\newblock In {\em Proceedings of the IEEE/CVF Conference on Computer Vision and
  Pattern Recognition}, pages 1141--1150, 2022.

\bibitem{schorr2021neuroscope}
Christian Schorr, Payman Goodarzi, Fei Chen, and Tim Dahmen.
\newblock Neuroscope: An explainable ai toolbox for semantic segmentation and
  image classification of convolutional neural nets.
\newblock {\em Applied Sciences}, 11(5):2199, 2021.

\bibitem{schramowski2020making}
Patrick Schramowski, Wolfgang Stammer, Stefano Teso, Anna Brugger, Franziska
  Herbert, Xiaoting Shao, Hans-Georg Luigs, Anne-Katrin Mahlein, and Kristian
  Kersting.
\newblock Making deep neural networks right for the right scientific reasons by
  interacting with their explanations.
\newblock {\em Nature Machine Intelligence}, 2(8):476--486, 2020.

\bibitem{schrouff2021best}
Jessica Schrouff, Sebastien Baur, Shaobo Hou, Diana Mincu, Eric Loreaux, Ralph
  Blanes, James Wexler, Alan Karthikesalingam, and Been Kim.
\newblock Best of both worlds: local and global explanations with
  human-understandable concepts.
\newblock {\em arXiv preprint arXiv:2106.08641}, 2021.

\bibitem{simonyan2014very}
Karen Simonyan and Andrew Zisserman.
\newblock Very deep convolutional networks for large-scale image recognition.
\newblock {\em arXiv preprint arXiv:1409.1556}, 2014.

\bibitem{stock2018convnets}
Pierre Stock and Moustapha Cisse.
\newblock Convnets and imagenet beyond accuracy: Understanding mistakes and
  uncovering biases.
\newblock In {\em Proceedings of the European Conference on Computer Vision
  (ECCV)}, pages 498--512, 2018.

\bibitem{tsunakawa2019contrastive}
Hideomi Tsunakawa, Yoshitaka Kameya, Hanju Lee, Yosuke Shinya, and Naoki
  Mitsumoto.
\newblock Contrastive relevance propagation for interpreting predictions by a
  single-shot object detector.
\newblock In {\em 2019 International Joint Conference on Neural Networks
  (IJCNN)}, pages 1--9. IEEE, 2019.

\bibitem{vielhaben2022sparse}
Johanna Vielhaben, Stefan Blücher, and Nils Strodthoff.
\newblock Sparse subspace clustering for concept discovery ({SSCCD}).
\newblock {\em arXiv preprint arXiv:2203.06043}, 2022.

\bibitem{vinogradova2020towards}
Kira Vinogradova, Alexandr Dibrov, and Gene Myers.
\newblock Towards interpretable semantic segmentation via gradient-weighted
  class activation mapping (student abstract).
\newblock In {\em Proceedings of the AAAI Conference on Artificial
  Intelligence}, volume~34, pages 13943--13944, 2020.

\bibitem{wan2020segnbdt}
Alvin Wan, Daniel Ho, Younjin Song, Henk Tillman, Sarah~Adel Bargal, and
  Joseph~E Gonzalez.
\newblock {SegNBDT}: Visual decision rules for segmentation.
\newblock {\em arXiv preprint arXiv:2006.06868}, 2020.

\bibitem{yamauchi2022spatial}
Toshinori Yamauchi and Masayoshi Ishikawa.
\newblock Spatial sensitive grad-cam: Visual explanations for object detection
  by incorporating spatial sensitivity.
\newblock In {\em 2022 IEEE International Conference on Image Processing
  (ICIP)}, pages 256--260. IEEE, 2022.

\bibitem{yan2022gsm}
Yicheng Yan, Xianfeng Li, Ying Zhan, Lianpeng Sun, and Jinjun Zhu.
\newblock Gsm-hm: Generation of saliency maps for black-box object detection
  model based on hierarchical masking.
\newblock {\em IEEE Access}, 10:98268--98277, 2022.

\bibitem{zhou2015object}
Bolei Zhou, Aditya Khosla, {\`{A}}gata Lapedriza, Aude Oliva, and Antonio
  Torralba.
\newblock Object detectors emerge in deep scene {CNNs}.
\newblock In {\em 3rd International Conference on Learning Representations
  (ICLR)}, 2015.

\bibitem{zhou2018Interpretable}
Bolei Zhou, Yiyou Sun, David Bau, and Antonio Torralba.
\newblock Interpretable basis decomposition for visual explanation.
\newblock In {\em Proceedings of the European Conference on Computer Vision
  (ECCV)}, pages 119--134, 2018.

\end{thebibliography}
}

\appendix
\newpage
\clearpage
\newpage
\setcounter{figure}{0}
\renewcommand\thefigure{\thesection.\arabic{figure}}  
\setcounter{equation}{0}
\renewcommand\theequation{\thesection.\arabic{equation}}  

\section{Appendix}
    In the appendix,
    we provide additional information and results complementing the main manuscript.
    Beginning with Section~\ref{app:technical_details},
    details about the models, datasets, as well as technical minutae on the computation of \gls{lrp} attributions are presented.
    Thereafter,
    Section~\ref{app:examples}
    provides additional examples of \gls{lcrp} explanations.
    This is followed by Sections~\ref{app:faithfulness} and \ref{app:complexity},
    where more results on the evaluation of explanations are given regarding faithfulness and complexity, respectively.
    Finally,
    in Section~\ref{app:context},
    additional results for the measurement of context scores of concepts are shown and discussed.

    \subsection{Technical Details}
    \label{app:technical_details}

    In the following, the models and datasets are presented, as well as details about the used \gls{lrp} rules and their implementation. 

    \paragraph{Models and Datasets}
    
        The UNet \cite{ronneberger2015u} model is implemented using the \texttt{segmentation\_models\_pytorch} framework \cite{Iakubovskii:2019} and consists of a VGG-13 \cite{simonyan2014very} encoder with BatchNorm layers, and weights pre-trained on ImageNet \cite{russakovsky2015imagenet}.
        The model is trained for 100 epochs with a batch size of 40 on the CityScapes \cite{cordts2016cityscapes} dataset with an initial learning rate of $10^{-3}$ using the Adam optimizer.
        The learning rate is reduced to $10^{-4}$ after 50 and $5\cdot 10^{-5}$ after 75 epochs.
        Images are resized to a height of 256 and width of 512 pixels
        and normalized using a mean of (0.485, 0.456, 0.406) and standard deviation of (0.229, 0.224, 0.225) over the three RGB color dimensions\footnote{as proposed at \url{https://pytorch.org/vision/stable/models.html}}.
        During training,
        we further apply random cropping to $256\times 256$ pixels, a random horizontal flip (50\,\% probability),
        brightness, saturation, hue and contrast perturbation (20\,\% strength, 50\,\% probability),
        as well as adding random Gaussian noise (zero mean, variance between 10 and 50, 50\,\% probability).
        The final training results in a pixel-wise accuracy of 74.7\,\% and a mean intersection over union score of 35.1\,\%.
    
        The DeepLabV3+ \cite{chen2018encoder} model is based on the PyTorch implementation \cite{fang2019deeplabv3plus} with a ResNet-50 \cite{he2016deep} backbone and pre-trained on the Pascal VOC 2012 dataset.
    
        The YOLOv6 model is based on the PyTorch implementation of the authors \cite{li2022yolov6} and corresponds to the ``small'' variant named YOLOv6s trained on the MS COCO 2017 dataset \cite{lin2014microsoft}. 
        
        Similarly,
        the YOLOv5 model is based on the PyTorch implementation of \cite{jocher2021yolov5} and corresponds to the ``medium'' variant named YOLOv5m trained on MS COCO 2017.
        
        \paragraph{LRP Canonization}
        Canonization procedures restructure a model into a functionally equivalent architecture to which established attribution \gls{lrp}-rules can be applied. 
        Canonization efforts typically concentrate on replacing the BatchNorm layer \cite{hui2019batchnorm, guillemot2020breaking} or handling skip connections~\cite{binder2020notes} as in the ResNet architecture \cite{he2016deep}.
        
        Regarding canonization, BatchNorm layers are merged into the preceding linear layer for all models. 
        This way, significant improvements in terms of explanations can be achieved \cite{motzkus2022measurably}.
    
        Further,
        in the \texttt{Bottleneck} modules of DeepLabV3+ (based on the ResNet) and YOLOv5, as well as the \texttt{RepVGGBlock} module (based on \cite{ding2021repvgg}) of the YOLOv6, 
        activations of different parallel operations are merged using a sum operation.
        Here,
        a \texttt{Sum} module is inserted to handle the summation,
        and to which a \texttt{Norm}-rule can be applied to normalize relevances during the relevance backpropagation step, as available in \cite{anders2021software} and discussed in \cite{binder2020notes}.

        The authors of \cite{ding2021repvgg} also present a canonization procedure for the \texttt{RepVGGBlock} module of YOLOv6, which merges all linear operations of the module into a single convolutional layer,
        which we apply to layer \texttt{model.backbone.stem}.
        
        \paragraph{LRP Rules}
        Multiple rules for \gls{lrp} were defined in the literature \cite{kohlbrenner2020towards}. 
        The \gls{lrp}-$\varepsilon$ rule is based on the basic decomposition rule (see Equation~\eqref{eq:lrp_basic_decomp}) and is given as
        \begin{align}
        R^{(l,\: l+1)}_{i \leftarrow j} & =  \frac{z_{ij}}{\preact_j + \varepsilon \cdot \text{sign}(\preact_j)}R_j^{l+1}
        \end{align}
        with $z_{ij}$ describing the contribution of neuron $i$ to the activation of neuron $j$,
        and aggregated pre-activations $z_j = \sum_i z_{ij}$.
        The \gls{lrp}-$\varepsilon$ rule ensures that each neuron receives the attribution value (fraction), that it contributed to the output. 
        The added parameter $\varepsilon \in \mathbb{R}^+$ stabilizes the division of the denominator, and ``dampens'' contradicting contributions for $|z_{ij}|\gg |z_j|$, as discussed in \cite{montavon2019layer}.
        However,
        $\varepsilon$ is usually set to a small value, \eg $10^{-6}$ in \cite{anders2021software},
        resulting in noisy attributions for deep networks due to large relevance values from contradicting contributions $|z_{ij}|\gg |z_j|$.
        In classification networks,
        the \gls{lrp}-$\varepsilon$ rule is often chosen for the dense layers,
        whereas increasingly stabilizing rules (presented in the following) are chosen for convolutional layers.
        In the YOLOv6 model implementation, 
        single BatchNorm layers with label \texttt{rbr\_identity} in \texttt{RepVGGBlock} modules exist without neighbouring linear layers.
        We apply the \gls{lrp}-$\varepsilon$ rule for these individual BatchNorm layers.
    
        Alternatively,
        the \gls{lrp}-$\gamma$ rule is defined as
        \begin{align} \label{eq:appendix:gamma-rule}
        R^{(l,\: l+1)}_{i \leftarrow j} & =  
        \frac{z_{ij} + \gamma z_{ij}^+}
        {z_j + \gamma \sum_i z_{ij}^+}
        R_j^{l+1}
        \end{align}
        with positive parameter $\gamma \in \mathbb{R}^+$ and $(\cdot)^+ =\max(0, \cdot)$.
        The function of the \gls{lrp}-$\gamma$ rule is to favor positive contributions and at the same 
        limit the unbounded growth potential of positive and negative relevance in the backpropagation step. 
        Note, 
        that the term $\gamma \sum_i z_{ij}^+$ (for strictly positive $z_j$) effectively corresponds to the $\varepsilon$ in Equation~\eqref{eq:lrp_basic_decomp},
        but scaled to the magnitude of contributions $z_{ij}^+$.
        The \gls{lrp}-$\gamma$ rule has thus been found to be effective in reducing noisy attributions,
        and the amount of ``filtering'' can be controlled by one parameter $\gamma$,
        which is by default set to $\gamma = 0.25$ in the \text{zennit} framework \cite{anders2021software}.
        However,
        if $z_j<0$, 
        the denominator can become numerically unstable if $|z_j| \approx \gamma \sum_i z_{ij}^+$.
        Therefore,
        we use the generalized version presented in \cite{andeol2021learning}, and either favor positive contributions for $z_j>0$ or negative contributions for $z_j<0$, depending on the sign of $z_j$:
        \begin{align} \label{eq:appendix:gamma-rule-general}
            R^{(l,\: l+1)}_{i \leftarrow j} & = 
            \begin{cases}
                    \frac{z_{ij} + \gamma z_{ij}^+}
            {z_j + \gamma \sum_i z_{ij}^+}
            R_j^{l+1} & \text{if } z_j>0  \\ 
                    \frac{z_{ij} + \gamma z_{ij}^-}
            {z_j + \gamma \sum_i z_{ij}^-}
            R_j^{l+1} & \text{else}
            \end{cases}
            ~.
        \end{align}
        
        The \gls{lrp}-$z^+$ rule represents a third rule,
        defined as
        \begin{align} \label{eq:appendix:zplus-rule}
        R^{(l,\: l+1)}_{i \leftarrow j} & =  
        \frac{z_{ij}^+}
        {\sum_i z_{ij}^+}
        R_j^{l+1}
        \end{align}
        by only taking into account positive contributions $z_{ij}^+$.
        The \gls{lrp}-$z^+$ rule can be seen as the most stable or least noisy attribution method,
        representing Equation~\ref{eq:appendix:gamma-rule} with $\gamma \rightarrow \infty$.
        To receive explanations with high human interpretability and low amounts of noise,
        explanations are thus computed with the \gls{lrp}-$z^+$ rule applied to all convolutional layers of the models.
        For the YOLOv6 an exception is made for visual examples,
        as the \gls{lrp}-$\gamma$ rule also provides stable heatmaps,
        showing higher faithfulness at the same time, as discussed in \ref{sec:explanation_quality}.
        
        Finally, it is established practice to apply the \gls{lrp}-${\flat}$ rule to the first layer \cite{kohlbrenner2020towards, montavon2019layer} in order for the attributions to become invariant against normalizations applied in inpu space and making them more human-readable by minimally reducing heatmap fidelity. 
        The LRP-$\flat$ rule is hereby defined as
        \begin{align} 
        R^{(l,\: l+1)}_{i \leftarrow j} & =  \frac{1}{\sum_i 1}R_j^{l+1}\,.
        \end{align}
        Using the \gls{lrp}-${\flat}$ rule, the relevance of upper-level neuron $j$ is equally distributed to all connected lower-level neurons disregarding any influence of learned weights or input features. 
        For all models, the \gls{lrp}-$\flat$ rule is applied to the first convolutional layer to smooth the attribution map in input space and yield robust concept localization. 
        
        \paragraph{Software}
        \gls{lcrp} as an extension to \gls{crp} is implemented based on the open-source \gls{crp} framework \texttt{zennit-crp}\footnote{Available at the GitHub repository \url{https://github.com/rachtibat/zennit-crp}.} \cite{achtibat2022towards} toolbox for PyTorch and \gls{lrp} framework \texttt{zennit} \cite{anders2021software}.

    \subsection{Explanation Examples Using L-CRP}
    \label{app:examples}
        In the following,
        two additional \gls{lcrp} explanation examples are shown,
        which extend the \gls{lrp} heatmaps shown in Figure~\ref{fig:local_to_glocal_seg}.
    
        \begin{figure} 
                \centering
                     
                    \includegraphics[width=1\linewidth]{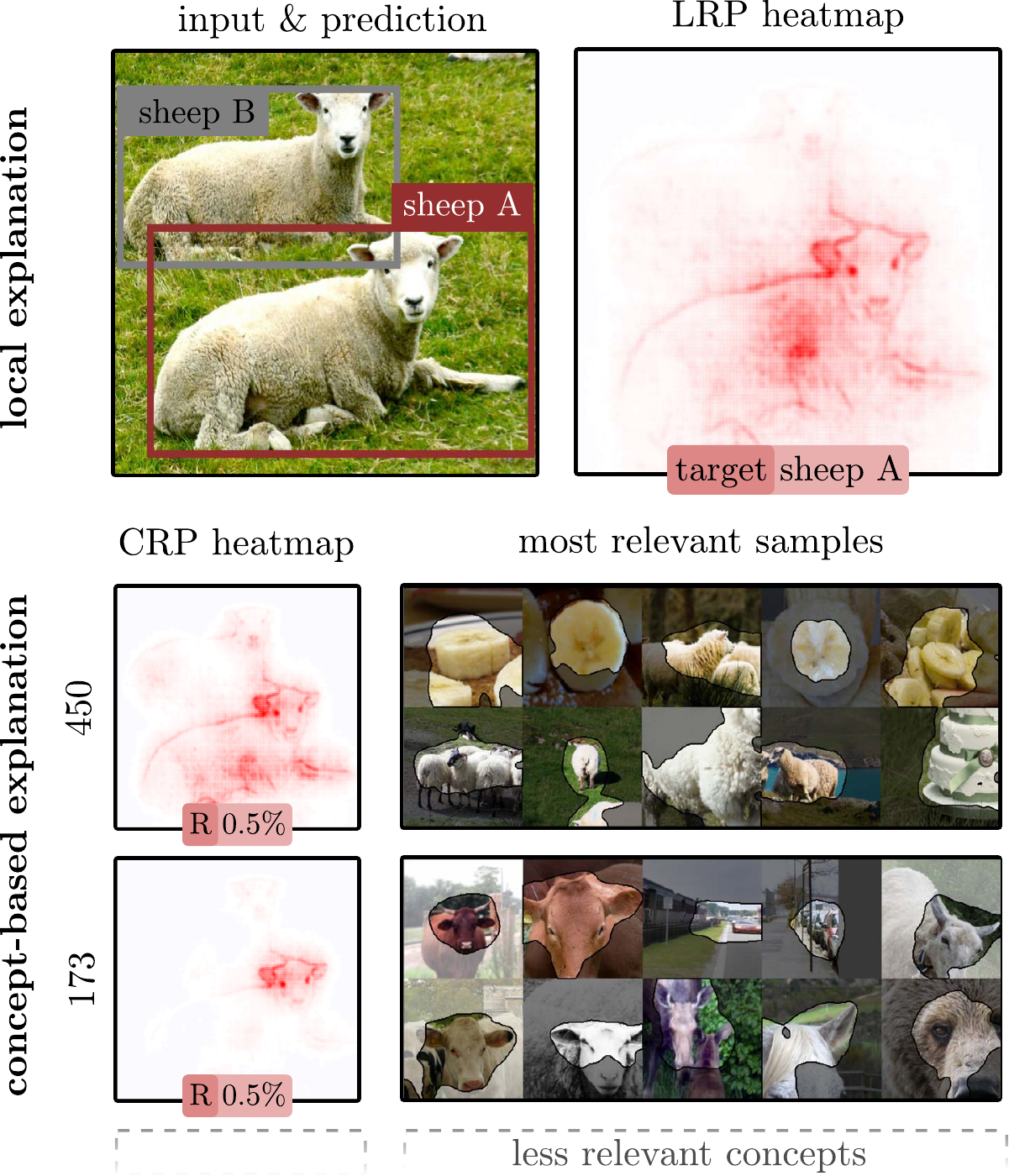}
                    \captionof{figure}{Concept-based explanation with \gls{lcrp} for object detection using the YOLOv5 model. (\emph{Top}): \gls{lrp} heatmap for detection of a sheep. The heatmap marks the center of the body as well as head. (\emph{Bottom}): Inspecting the two most relevance concepts in layer \texttt{7.conv} using \gls{lcrp} shows,
                    that the model perceives the bright fur texture (concept 450) as well as the frontal face with ears protruding to either side (concept 173).}
                    \label{fig:appendix:local_to_glocal_obdet}
            \end{figure}
    
        The first example shown in Figure~\ref{fig:appendix:local_to_glocal_obdet} corresponds to the sheep detection of the YOLOv5 model in Figure~\ref{fig:local_to_glocal_seg}.
        Here,
        the traditional \gls{lrp} heatmap indicates that parts of the sheep are relevant,
        such as the head or the fur part in the center of the bounding box.
        By applying \gls{lcrp} to layer \texttt{layer4.0.conv3},
        we achieve an understanding of what exactly is the model using in terms of concepts here.
        By analyzing the top-2 most relevant concepts,
        we learn that the model perceives, \eg, the white fur pattern of the sheep (concept 450) or the frontal face with ears protruding on both sides (concept 173).

        \begin{figure} 
            \centering
                \includegraphics[width=1\linewidth]{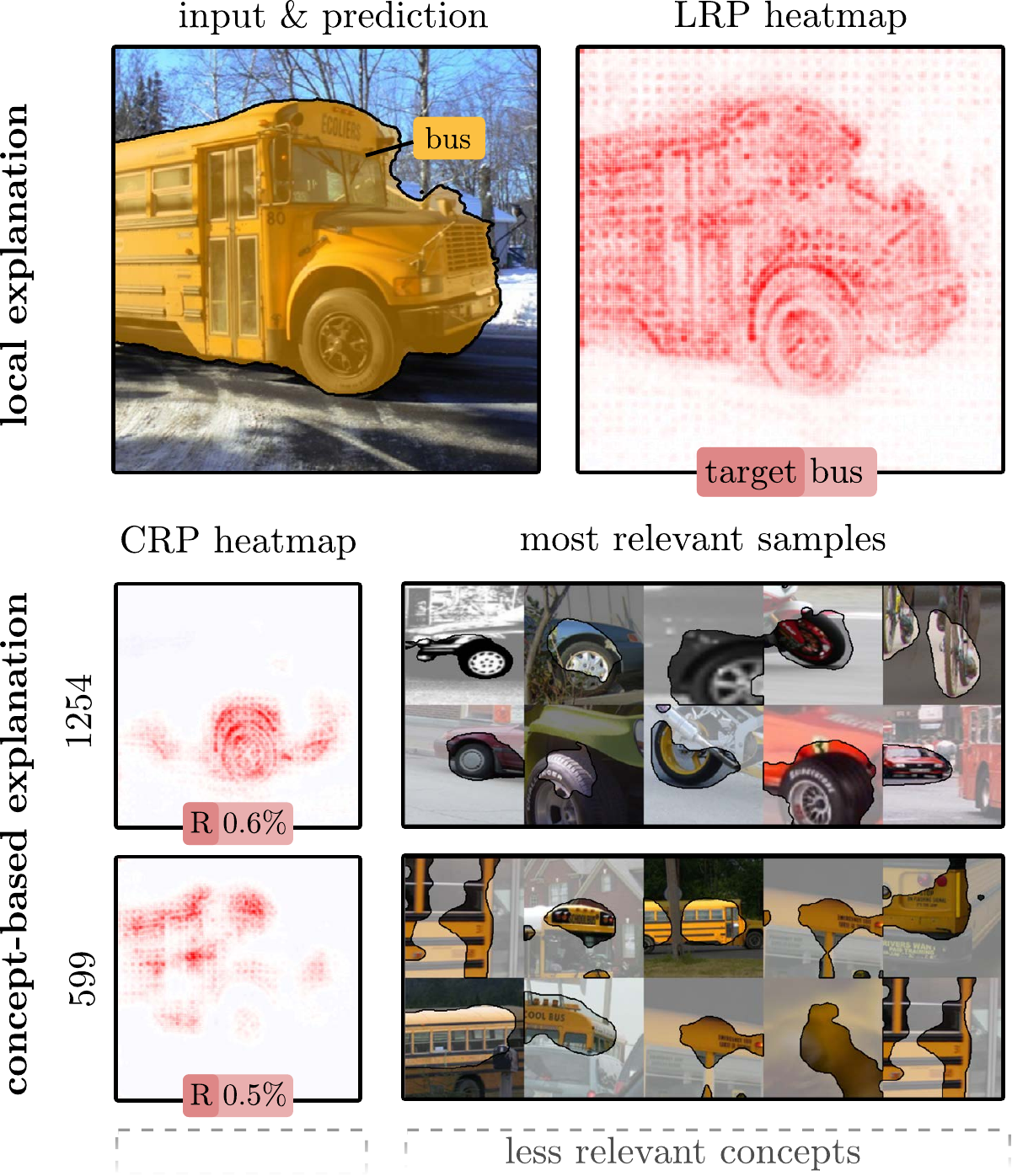}
                \captionof{figure}{Concept-based explanation with \gls{lcrp} for semantic segmentation using the DeepLabV3+ model. (\emph{Top}): \gls{lrp} heatmap for segmentation of a bus. The heatmap strongly resembles the segmentation output and marks the whole bus. (\emph{Bottom}): Inspecting the two most relevance concepts in layer \texttt{layer4.0.conv3} using \gls{lcrp} shows,
                that the model perceives the bus wheels (concept 1254) as well as the typical school bus color combination of yellow and black (concept 599).}
                \label{fig:appendix:local_to_glocal_semmseg}
        \end{figure}
    
        The second example shown in Figure~\ref{fig:appendix:local_to_glocal_semmseg} corresponds to the bus segmentation of the DeepLabV3+ model in Figure~\ref{fig:local_to_glocal_seg}.
        Here,
        the traditional \gls{lrp} heatmap indicates that all parts of the bus are relevant,
        strongly resembling the predicted segmentation itself.
        By applying \gls{lcrp} to layer \texttt{layer4.0.conv3},
        we achieve an understanding of what exactly is the model using in terms of concepts.
        Investigating the top-2 most relevant concepts,
        shows that the model perceives, \eg, the wheels of the bus (concept 1254) or the school bus color combination of yellow with black (concept 599).
    
    \subsection{Faithfulness}
    \label{app:faithfulness}
    In Section~\ref{sec:experiments:faithfulness},
    we propose to measure the faithfulness of concept attributions using various approaches based on relevance or activation.
    In this section,
    more detailed results for the discussion in Section~\ref{sec:experiments:faithfulness} are presented.
    \label{app:faithfulness}
        \begin{figure*} 
        \centering
            \includegraphics[width=.49\linewidth]{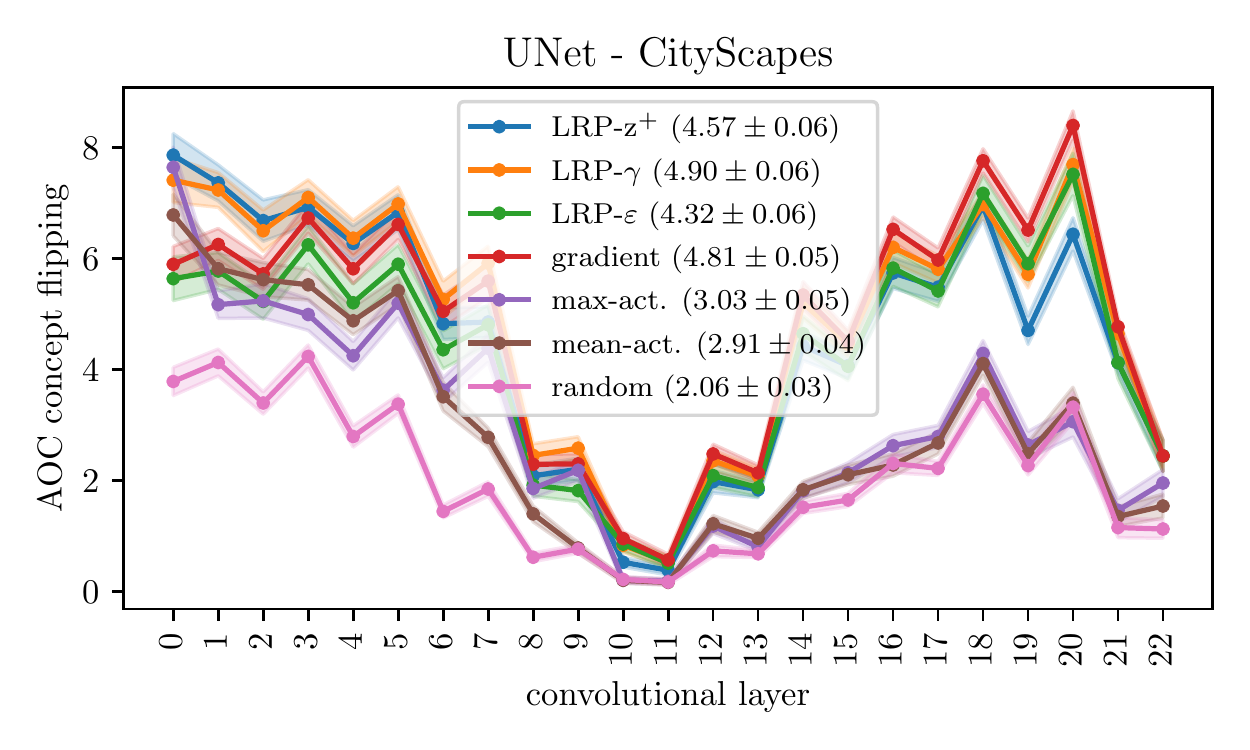}
            \includegraphics[width=.49\linewidth]{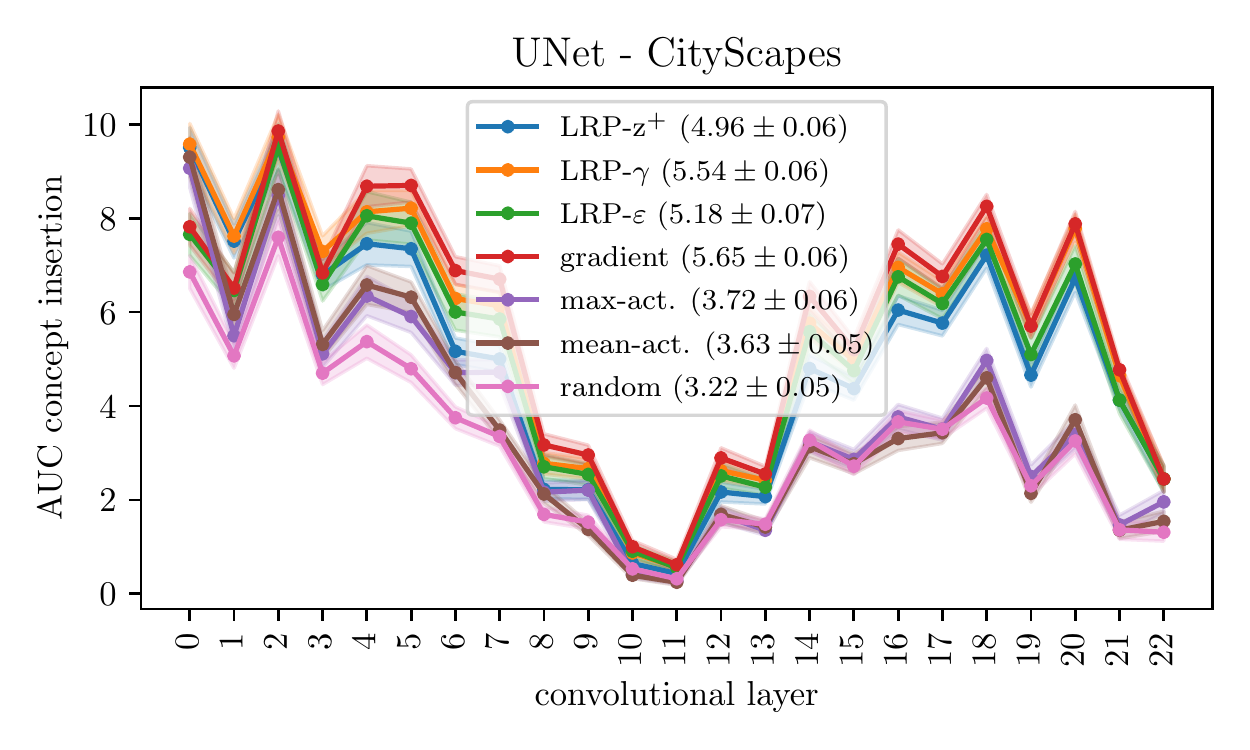}
            \includegraphics[width=.49\linewidth]{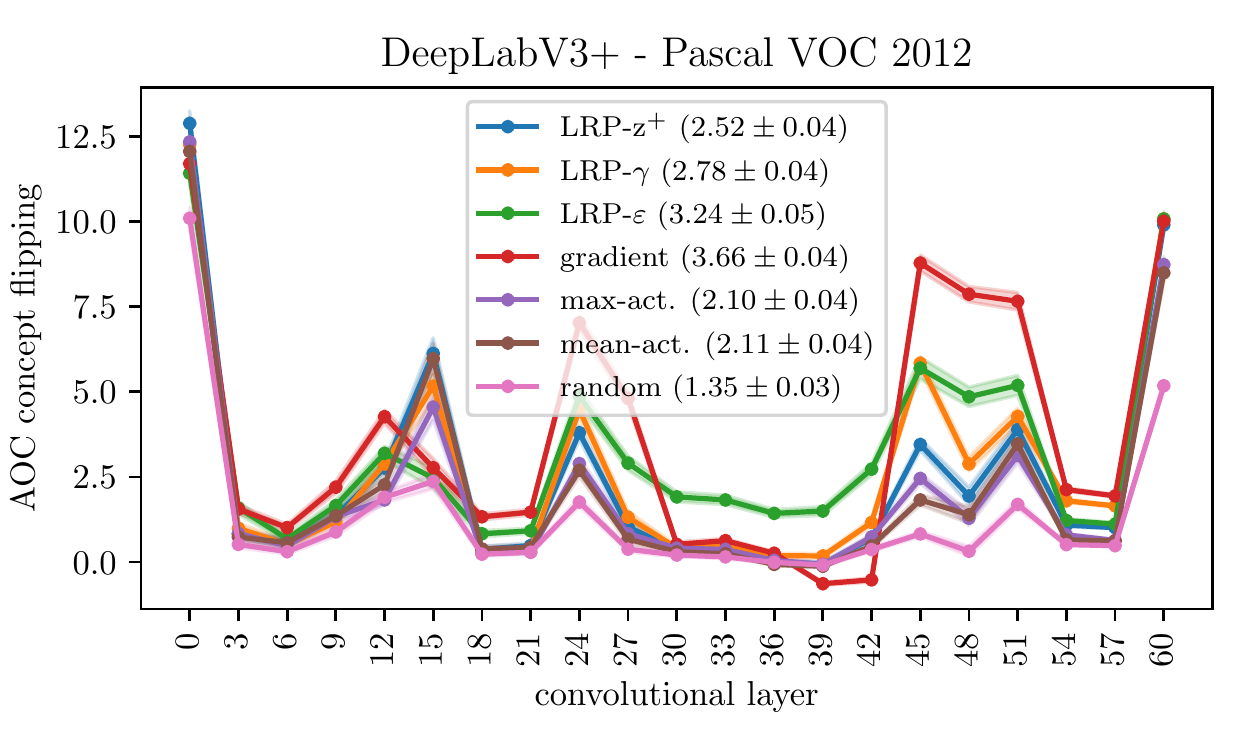}
            \includegraphics[width=.49\linewidth]{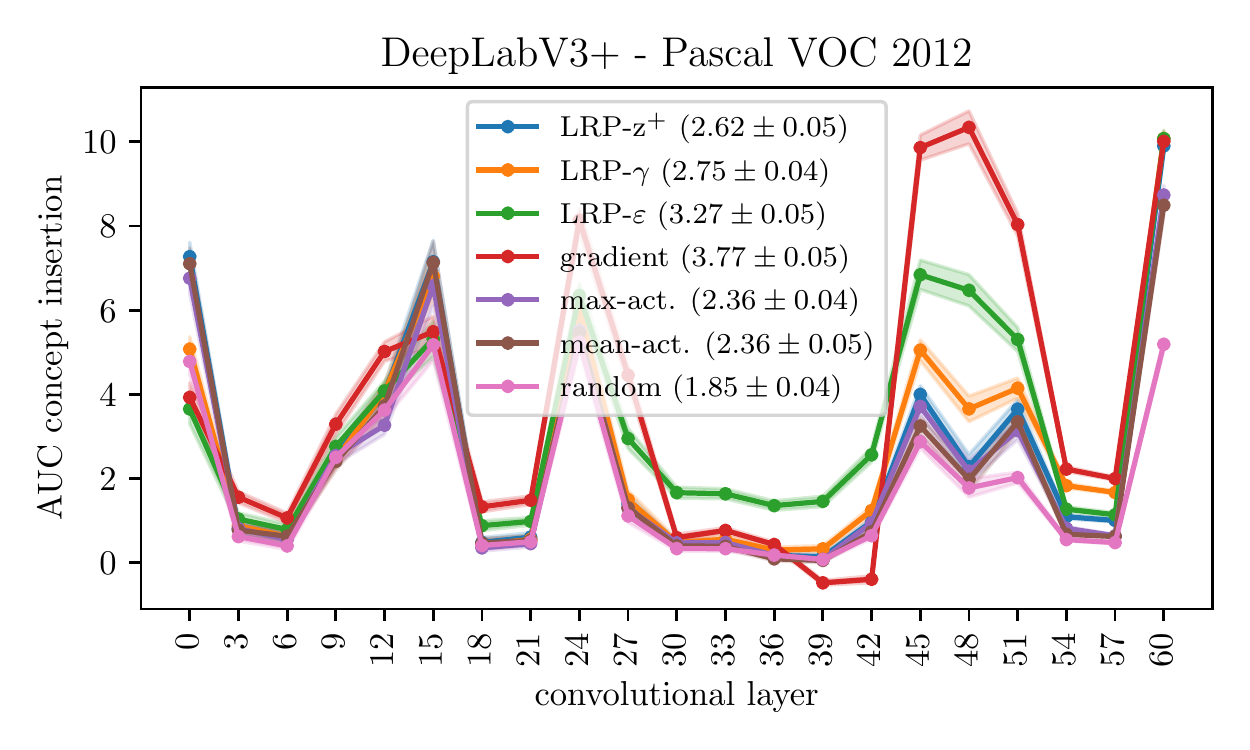}
            \includegraphics[width=.49\linewidth]{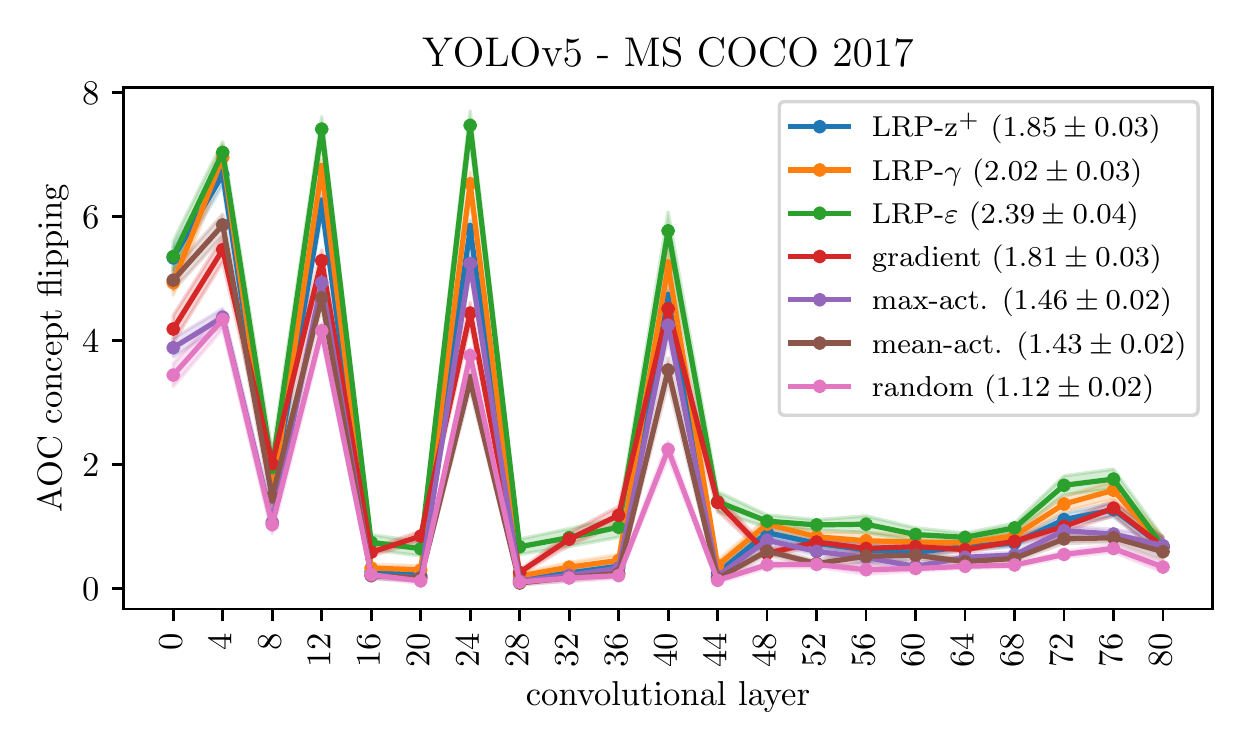}
            \includegraphics[width=.49\linewidth]{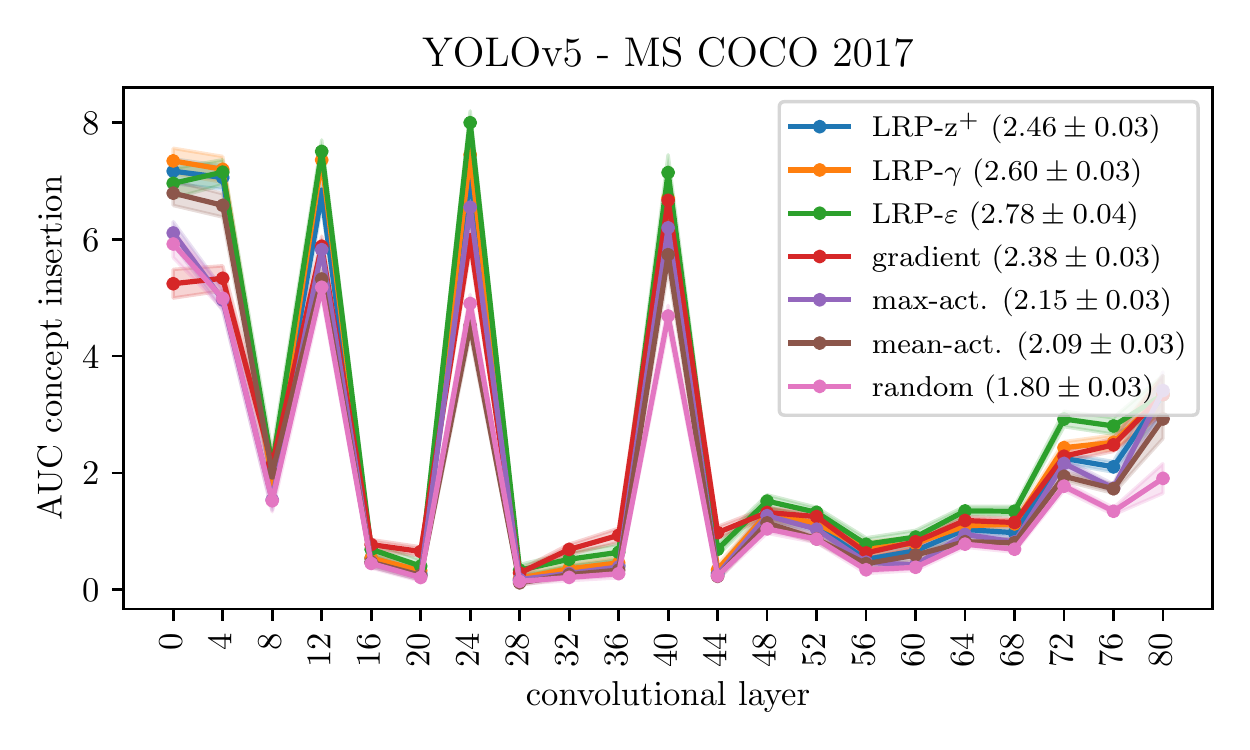}
            \includegraphics[width=.49\linewidth]{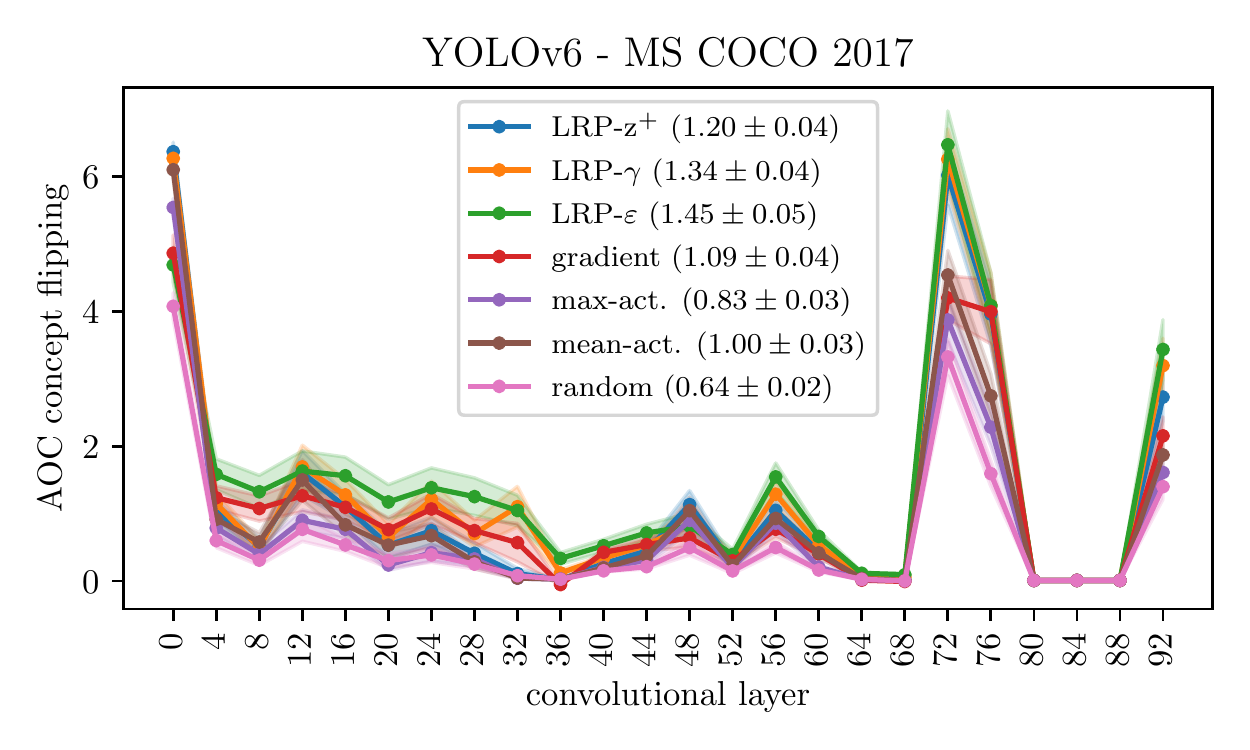}
            \includegraphics[width=.49\linewidth]{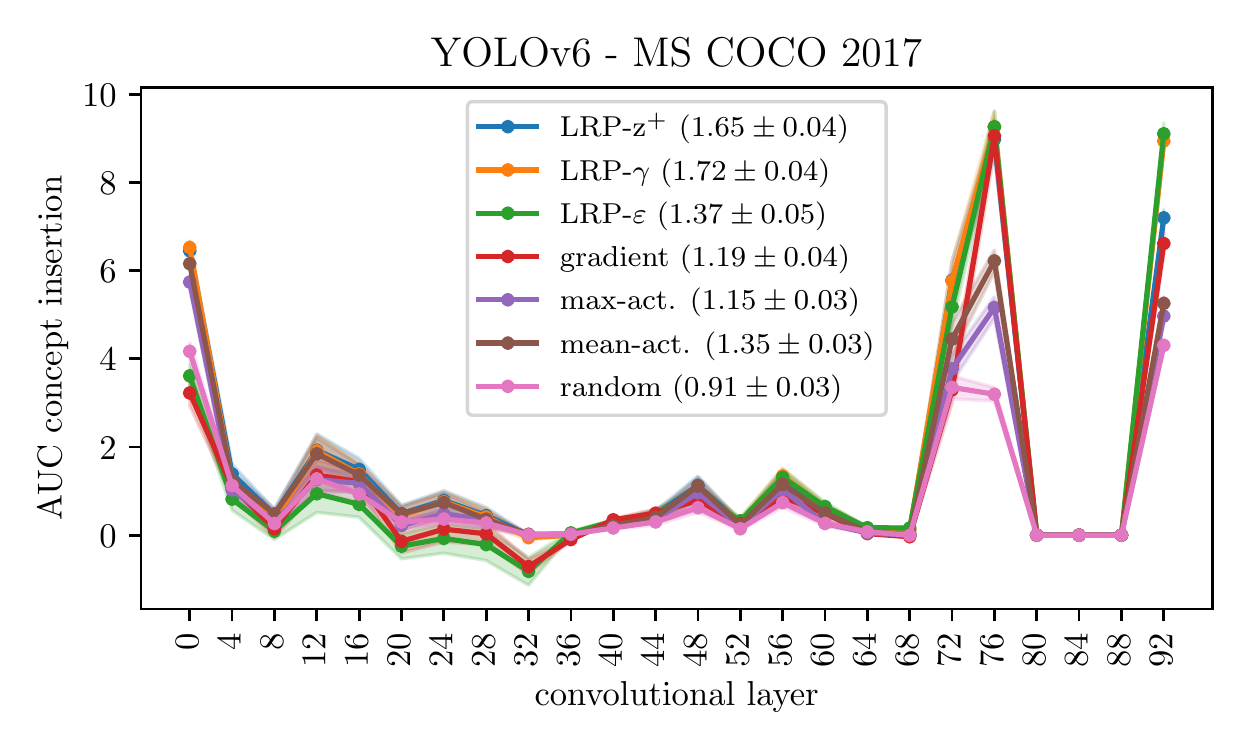}
            \captionof{figure}{Evaluating the faithfulness of concept attributions. (\emph{Left}): Concept flipping experiment. (\emph{Right}): Concept insertion experiment. The higher the \gls{aoc} or \gls{auc}, the more faithful the attribution method.
            The faithfulness scores over all layers is given in parenthesis.}
            \label{fig:appendix:faithfulness}
    \end{figure*}
    Specifically,
    we show additional results for the concept flipping and concept insertion experiment for different layers of the UNet,
    DeepLabV3+, YOLOv5 and YOLOv6 model in Figure~\ref{fig:appendix:faithfulness}.
    Here, every convolutional layer of the UNet,
    every second convolutional layer of the DeepLabV3+,
    and every fourth convolutional layer of the YOLO architectures is analyzed, 
    resulting in approximately 20 layers for each model.
    Here,
    the \gls{aoc} or \gls{auc} over the resulting curves are measured for 100 randomly chosen predictions, and the mean values are plotted with the \gls{sem} in semi-transparent color.
    The overall \gls{aoc} or \gls{auc} scores (of all layers) are given in parenthesis.

    Figure~\ref{fig:appendix:faithfulness} visualizes
    that the \gls{aoc} or \gls{auc} scores can vary strongly between layers of an attribution method.
    This is due to shortcut connections in the network,
    illustrating
    that some layers are used more strongly than others.
    Taking the UNet as an example,
    it shows that the further the layer in the encoder (up to layer 10),
    the lower the overall relevance.
    This indicates,
    that a large part of features is detected using the lower-level layers.

    \subsection{Complexity}

    In Section~\ref{sec:experiments:complexity},
    we measure and discuss the complexity of concept attribution scores for relevance and activation-based approaches by two means.
    Firstly,
    the standard deviation of class concept attribution scores is computed,
    and secondly,
    the amount of concepts needed to form 80\,\% of all attributions is measured.
    The lower the variation and the smaller the number of concepts to study,
    the easier it is to understand concept-based explanations.

    \label{app:complexity}
        \begin{figure*} 
        \centering
            \includegraphics[width=1\linewidth]{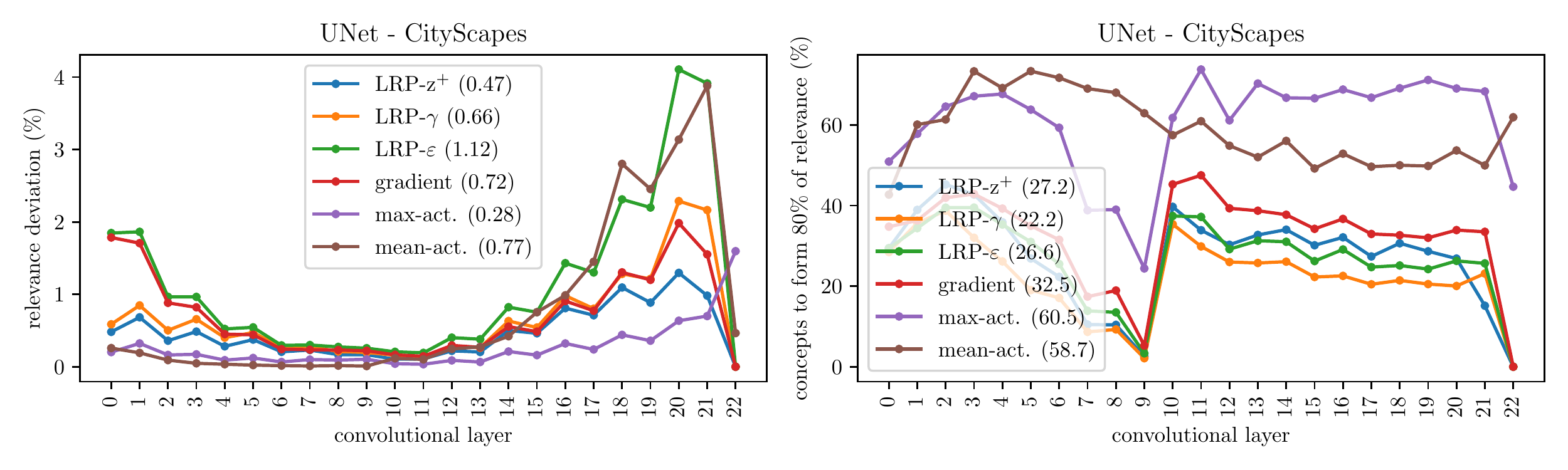}
            \includegraphics[width=1\linewidth]{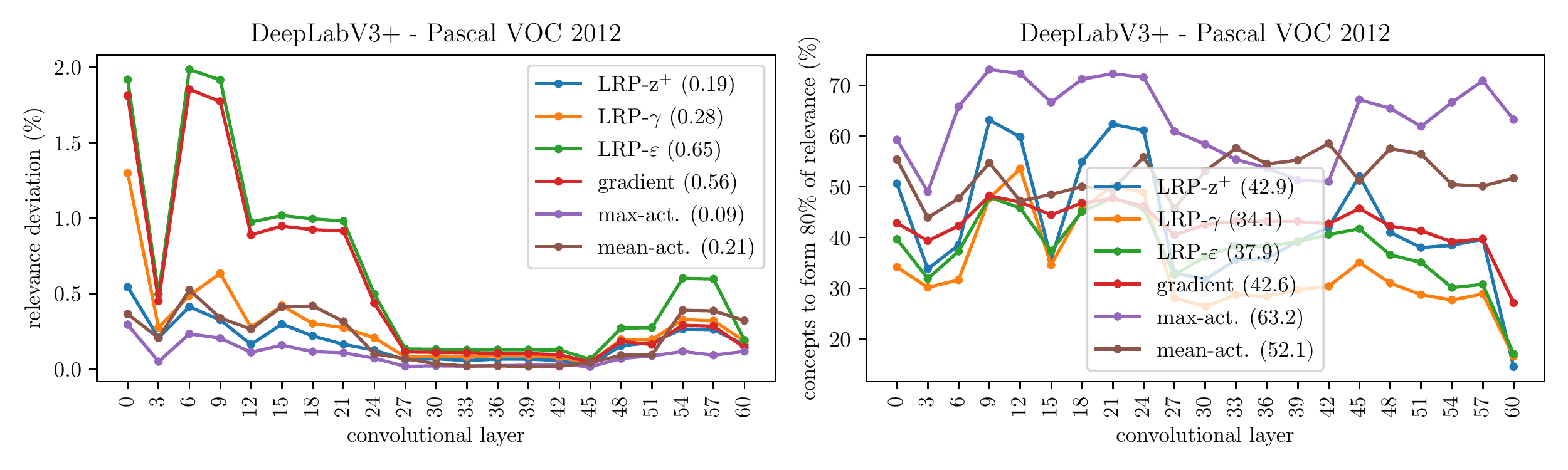}
            \includegraphics[width=1\linewidth]{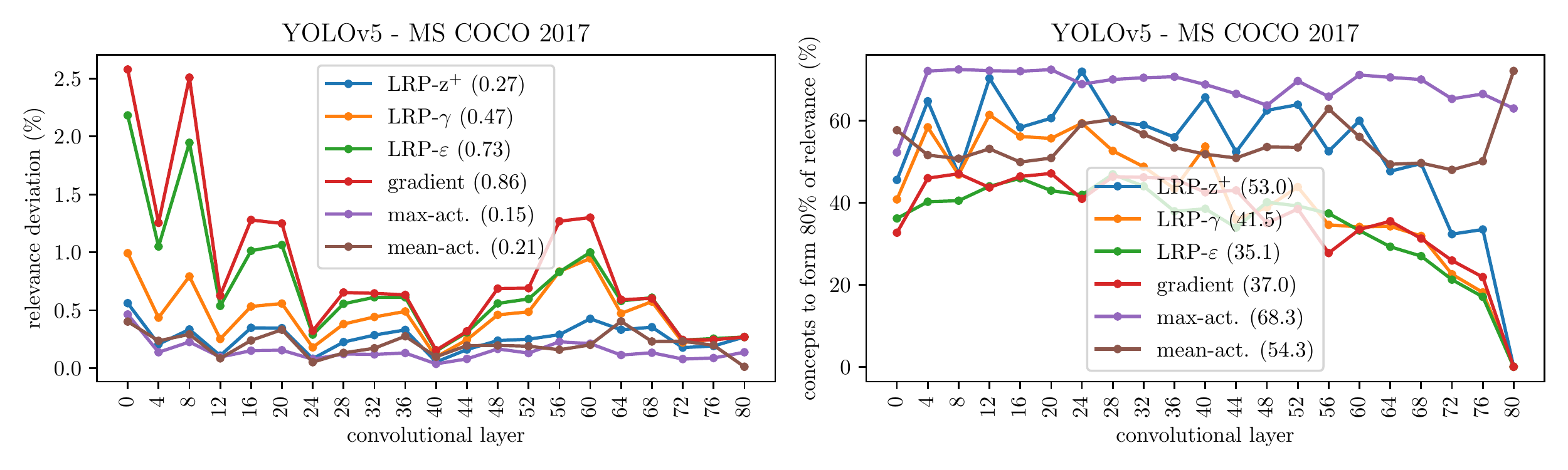}
            \includegraphics[width=1\linewidth]{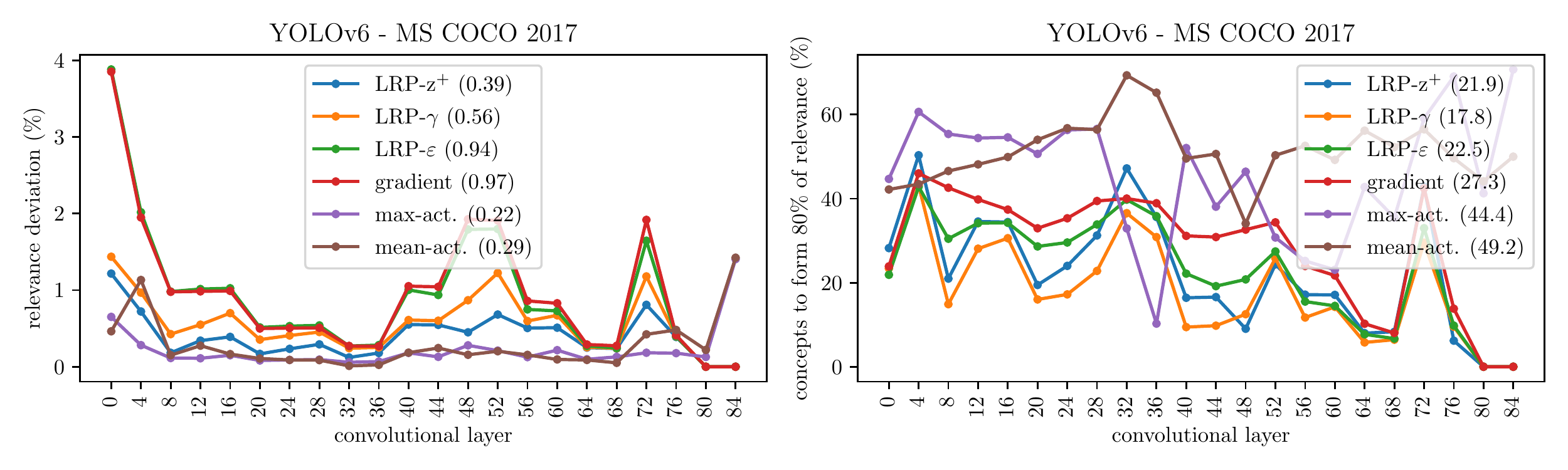}
            \captionof{figure}{Measuring the explanation complexity of concept attributions. (\emph{Left}): Mean standard deviation of relevances/attributions per class. The deviation over all layers is given in parentheses. (\emph{Right}): Number of concepts to form 80\,\% of all attributions. The concept number over all layers is given in parentheses. The lower the scores,
            the lower the complexity and the easier to read explanations.}
            \label{fig:appendix:complexity}
    \end{figure*}
    
    In the following,
    we present in Figure~\ref{fig:appendix:complexity} more detailed results of the complexity analysis discussed in Section~\ref{sec:experiments:complexity}.
    Here,
    the same layers of the models as in previous Section~\ref{app:faithfulness} are investigated.
    Regarding the analysis,
    concept attribution scores are collected over all predictions of the test dataset.
    Thereafter,
    in order to be interpretable as percentage scores,
    concept attributions are normalized to an absolute sum of one.
    
    In the first experiment,
    the mean standard deviation $\sigma_t$ of concept attribution scores $R_j(\x_i)$  for each class $t$ is measured as
    \begin{equation}
        \sigma_t =  \frac{1}{n_c} \sum_j^{n_c}  \sqrt{\frac{1}{n_s - 1} \sum_i^{n_s}  \left(R_j(\x_i) - \bar R_j\right)^2}
    \end{equation}
    with mean attribution $\bar R_j = \frac{1}{n_s} \sum_i^{n_s}  R_j(\x_i)$ over $n_s$ class samples and $n_c$ concepts.
    To form a final deviation score $\sigma$,
    the mean over all $n_t$ classes is computed as $\sigma = \frac{1}{n_t} \sum_t^{n_t} \sigma_t$.
    
    As shown in Figure~\ref{fig:appendix:complexity},
    it is visible,
    that especially in lower-level layers gradient and \gls{lrp}-$\varepsilon$ attributions tend to be noisy,
    which is expected,
    as this has also been observed for input-level heatmaps \cite{samek2021explaining}.
    However,
    gradient and \mbox{\gls{lrp}-$\varepsilon$} can also show noisy attributions in higher-level layers,
    \eg, in layer 20 of the UNet model.
    This indicates,
    that the commonly applied heuristic \cite{montavon2019layer} for \gls{lrp} heatmaps to use more faithful methods such as \gls{lrp}-$\varepsilon$ in higher-level layers,
    and more stable, but less faithful methods (\eg \gls{lrp}-z$^+$) in lower-level layer,
    might not always lead to stable concept-based explanations.

    In the second experiment,
    the fraction of concepts forming 80\,\% of attributions is calculated.
    Here,
    attributions are firstly sorted according to their magnitude in descending order.
    Thereafter,
    the cumulative distribution is computed,
    and the smallest number of channels computed for which the cumulative value is smaller than 80\,\%.
    Finally,
    the fraction is computed via division by the total number of channels.
    This value depends, \eg, on the feature specificity of concepts.
    For example,
    relevances are rather focused on a small set of concepts
    if they have a very specific function, compared to when all concepts are rather generic, leading to a uniform relevance distribution as all concepts are being used.

    \subsection{Context}
    
    In Section~\ref{sec:context},
    we compute context scores $C$ of concepts by measuring the number of concept attributions that correspond to the background of detected objects in the spatial dimension,
    as described by Equation~\eqref{eq:experiments:context:context_score}.
    We firstly evaluate the computed context scores comparing the use of latent activation maps, latent relevance maps, and \gls{lcrp} heatmaps,
    and thereafter show how to interact with the model and test its reliance on background features. For both parts,
    we in the following present additional information.

    \paragraph{Context Score Evaluation}
    The context scores are evaluated by computing background sensitivity scores $S$ for each concept via
    Equation~\eqref{eq:experiments:context:background_sensitivity}.
    The background sensitivity is hereby measured by perturbing the background of objects and tracking the change in concept attribution.

    Ideally,
    we expect that concepts with a high context score $C$ will also result in a high background sensitivity $S$.
    Therefore,
    we assume of faithful context scores to result in a high correlation $\rho$ as well as a small Root Mean Square Deviation (RMSD) value between context and background sensitivity scores.
    The correlation $\rho$ is calculated as
    \begin{equation}
        \rho = \sum_i \frac{(C_i - \bar C)(S_i - \bar S)}
        {\sqrt{\sum_j(C_j - \bar C)^2} \sqrt{\sum_k (S_k - \bar S)^2}}
    \end{equation}
    with means $\bar C = \frac{1}{m} \sum_i C_i$ and $\bar S = \frac{1}{m} \sum_i S_i$ over $m$ evaluated concepts.
    The RMSD is further given as
    \begin{equation}
        \text{RMSD} = \sqrt{\frac{1}{m} \sum_i (C_i - S_i)^2 }\,.
    \end{equation}
    The final correlation and RMSD values per model shown in Table~\ref{tab:experiments:context:evaluation} are computed by taking the mean over three layers.

    \label{app:context}
        \begin{figure*} 
        \centering
            \vspace{.5cm}
            \includegraphics[width=0.3\linewidth]{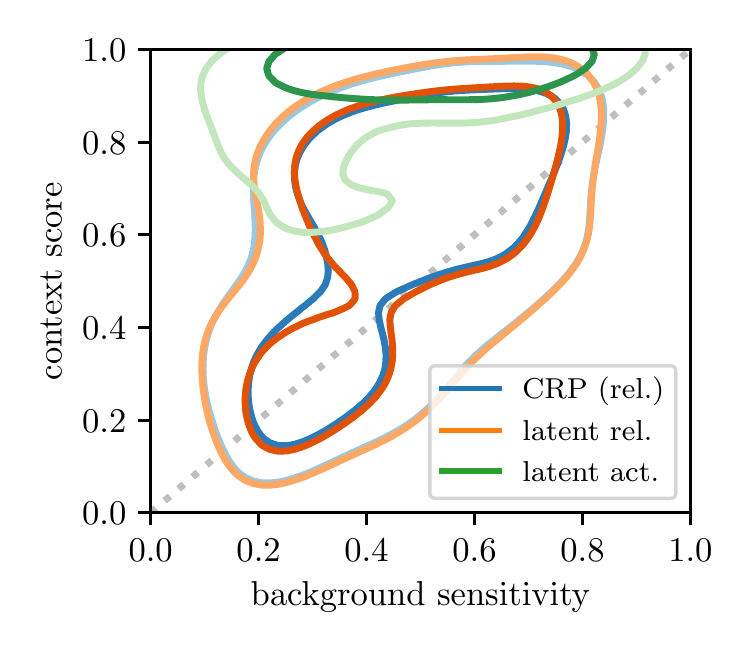}
            \includegraphics[width=0.3\linewidth]{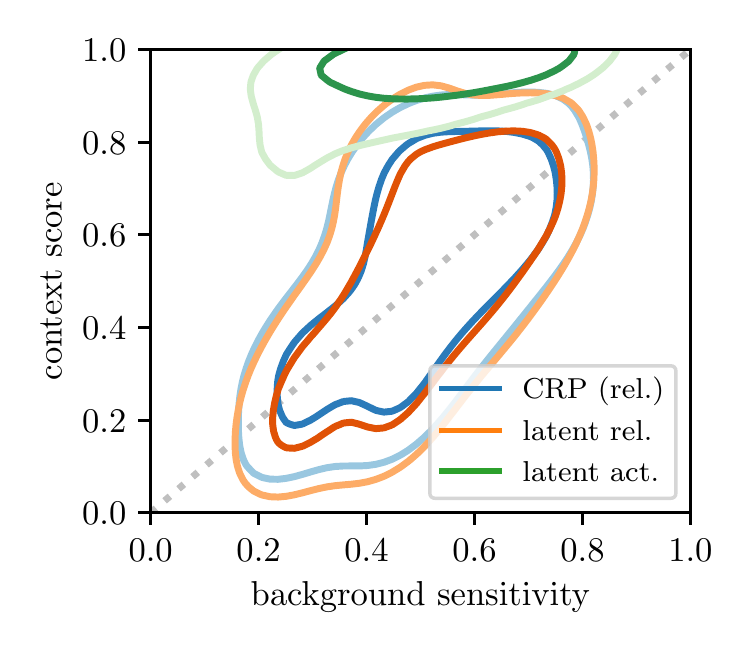}
            \includegraphics[width=0.3\linewidth]{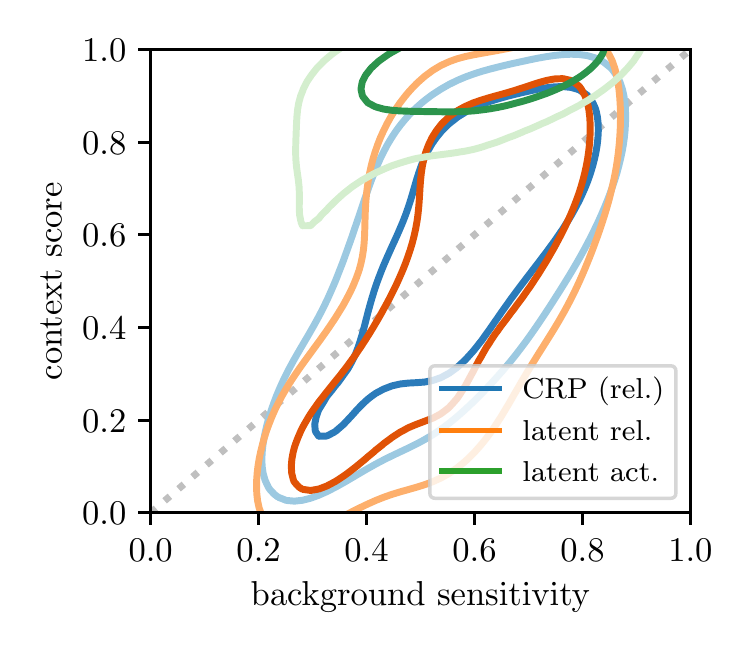}
            \includegraphics[width=0.3\linewidth]{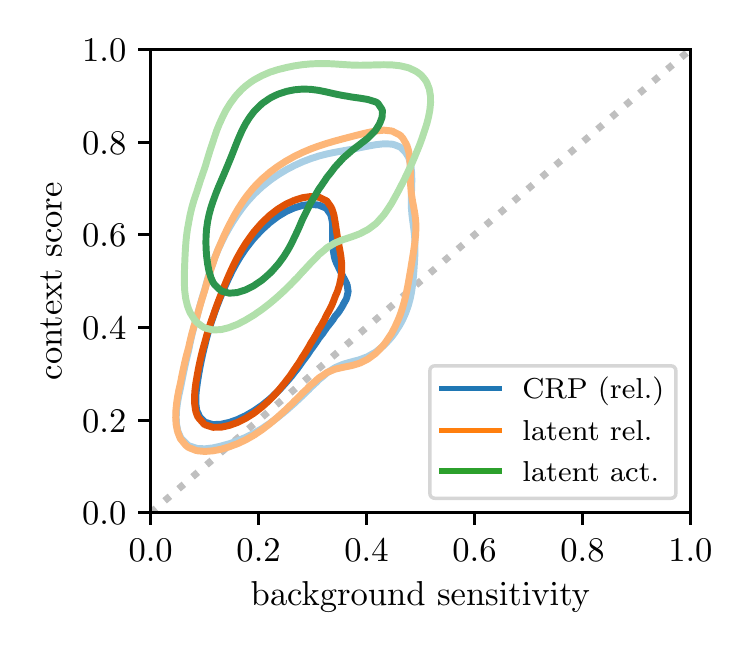}
            \includegraphics[width=0.3\linewidth]{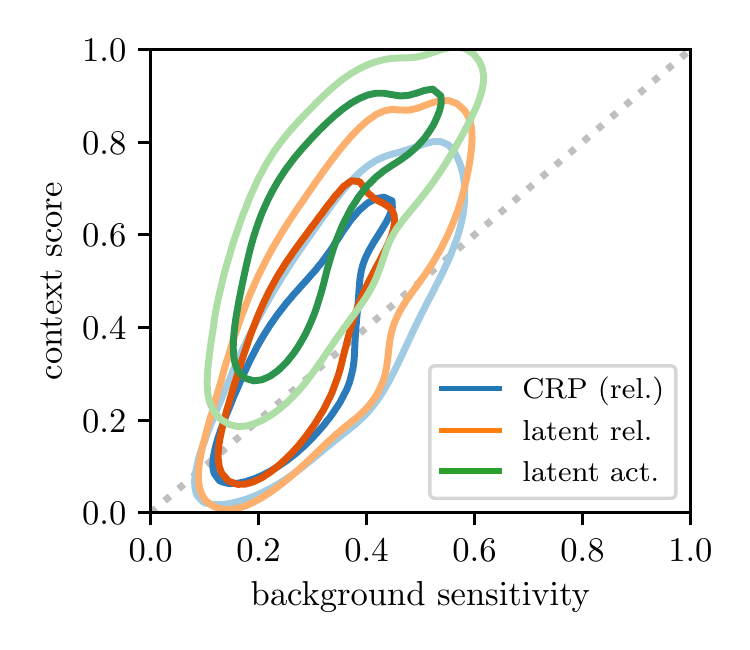}
            \includegraphics[width=0.3\linewidth]{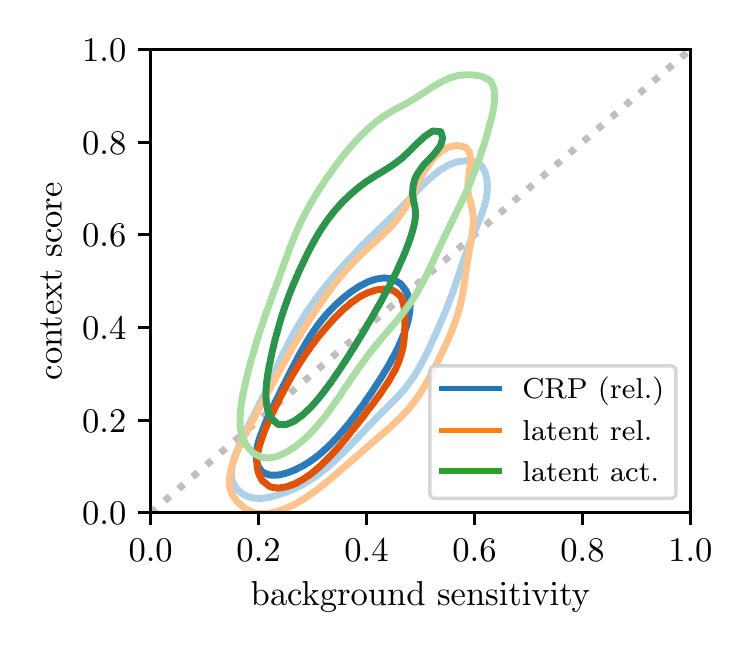}
            \includegraphics[width=0.3\linewidth]{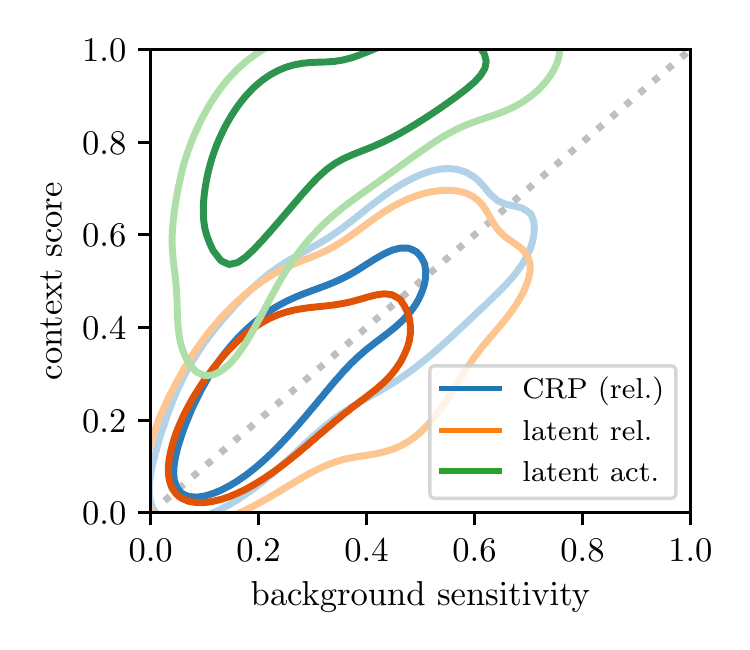}
            \includegraphics[width=0.3\linewidth]{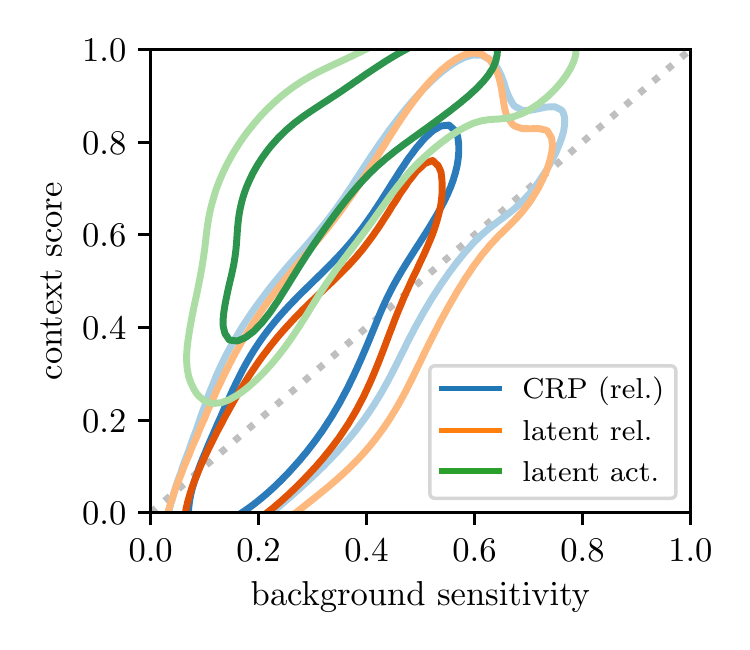}
            \includegraphics[width=0.3\linewidth]{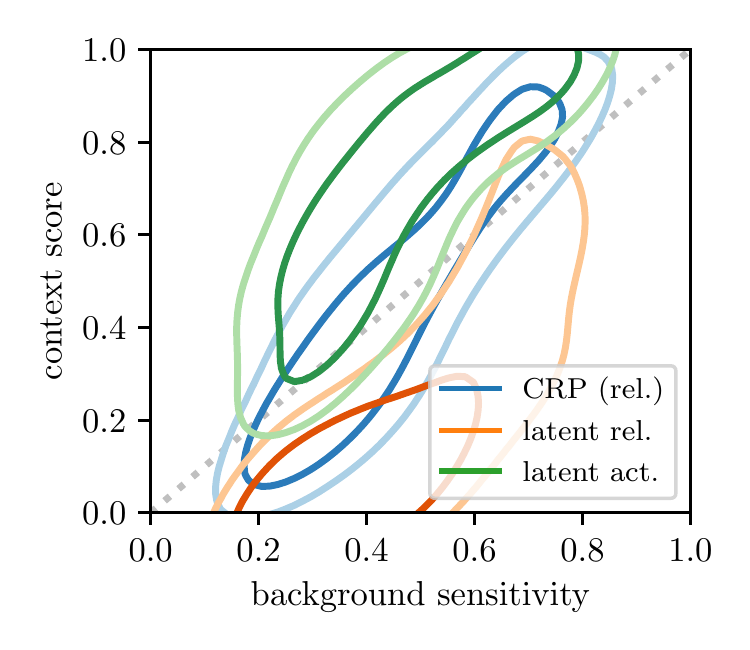}
            \includegraphics[width=0.3\linewidth]{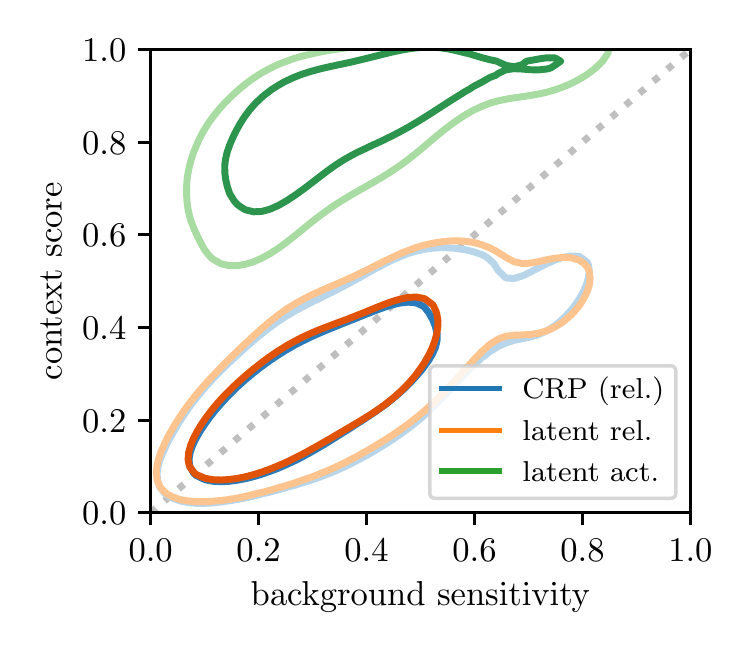}
            \includegraphics[width=0.3\linewidth]{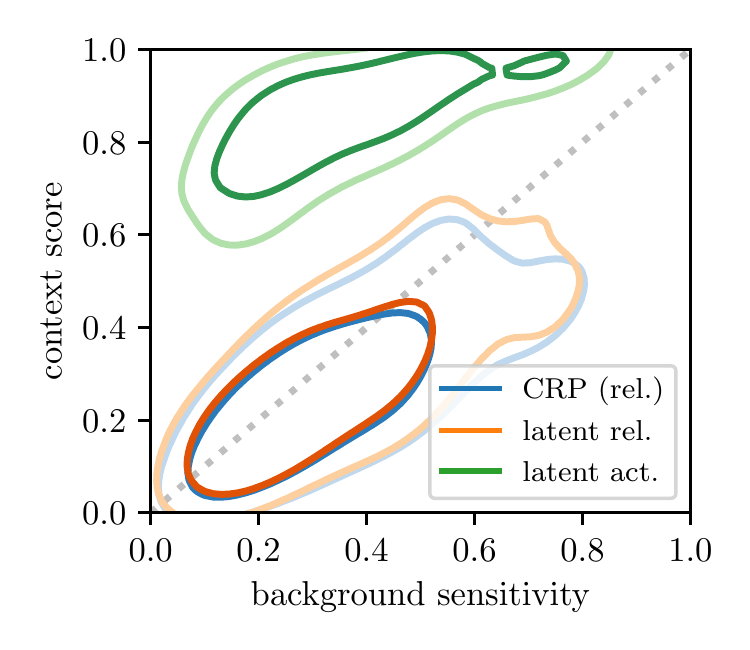}
            \includegraphics[width=0.3\linewidth]{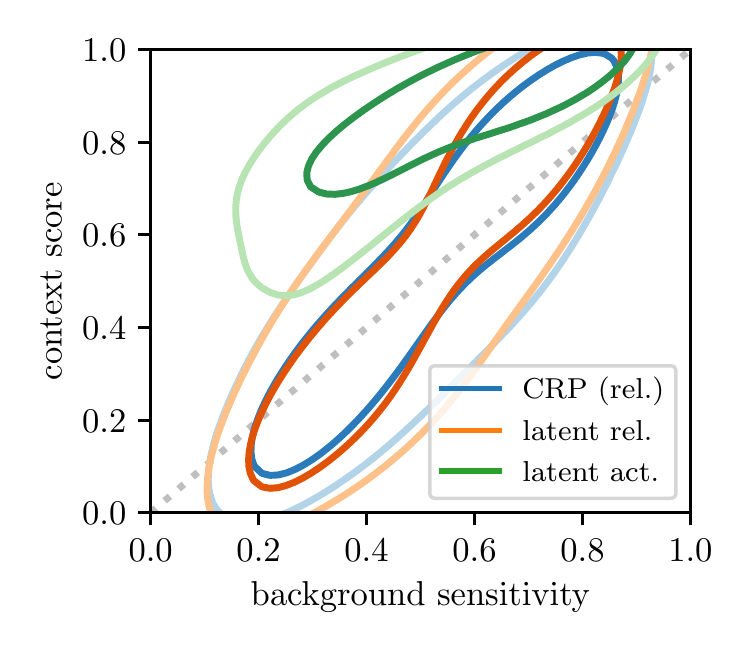}
            \vspace{.5cm}
            \captionof{figure}{Evaluation of the context scores $C$ of concepts by comparing these to background sensitivity scores $S$.
            (\emph{1st row}): Layers \texttt{features.0}, \texttt{features.5} and \texttt{features.10} (from left to right) of the UNet model.
            (\emph{2nd row}): Layers \texttt{layer3.0.conv1}, \texttt{layer4.0.conv3} and \texttt{layer4.2.conv2} (from left to right) of the DeepLabV3+ model.
            (\emph{3rd row}): Layers \texttt{6.cv3}, \texttt{8.cv3} and \texttt{10} (from left to right) of the YOLOv5 model.
            (\emph{4th row}): Layers \texttt{ERBlock\_3.0.rbr\_dense}, \texttt{ERBlock\_4.0.rbr\_dense} and \texttt{ERBlock\_5.0.rbr\_dense} (from left to right) of the YOLOv6 model.
            Shown are contours corresponding to 50\,\% and 80\,\% of values using estimated densities via Gaussian kernels of bandwidth 0.4.
            Ideally, a linear relationship exists as indicated by a dotted gray line.
            }
            \label{fig:appendix:complexity}
    \end{figure*}
    For a visualization of the resulting similarity between context scores and background sensitivity values,
    distribution plots are shown in Figure~\ref{fig:appendix:complexity}.
    Here,
    it becomes apparent,
    that using latent activation leads to an over-estimation of context scores,
    as they are often significantly higher than the corresponding background sensitivity values,
    as, \eg, for the three layers of the UNet shown in Figure~\ref{fig:appendix:complexity} (\emph{1st row}).
    It can also be seen,
    that the higher-level layers show the best alignment between $C$ and $S$ values (\emph{3rd column}),
    as features are more specialized,
    becoming better to characterize as either background or foreground concepts.
    
    \paragraph{Context-based Model Interaction}

    In the background concept flipping experiments in Section~\ref{sec:context},
    we interact with the model based on identified (by the model used) background concepts.
    Therefore,
    three background concepts are flipped successively and the change in the predicted object logit measured.
    
    \begin{figure*} 
        \centering
            \includegraphics[width=0.86\linewidth]{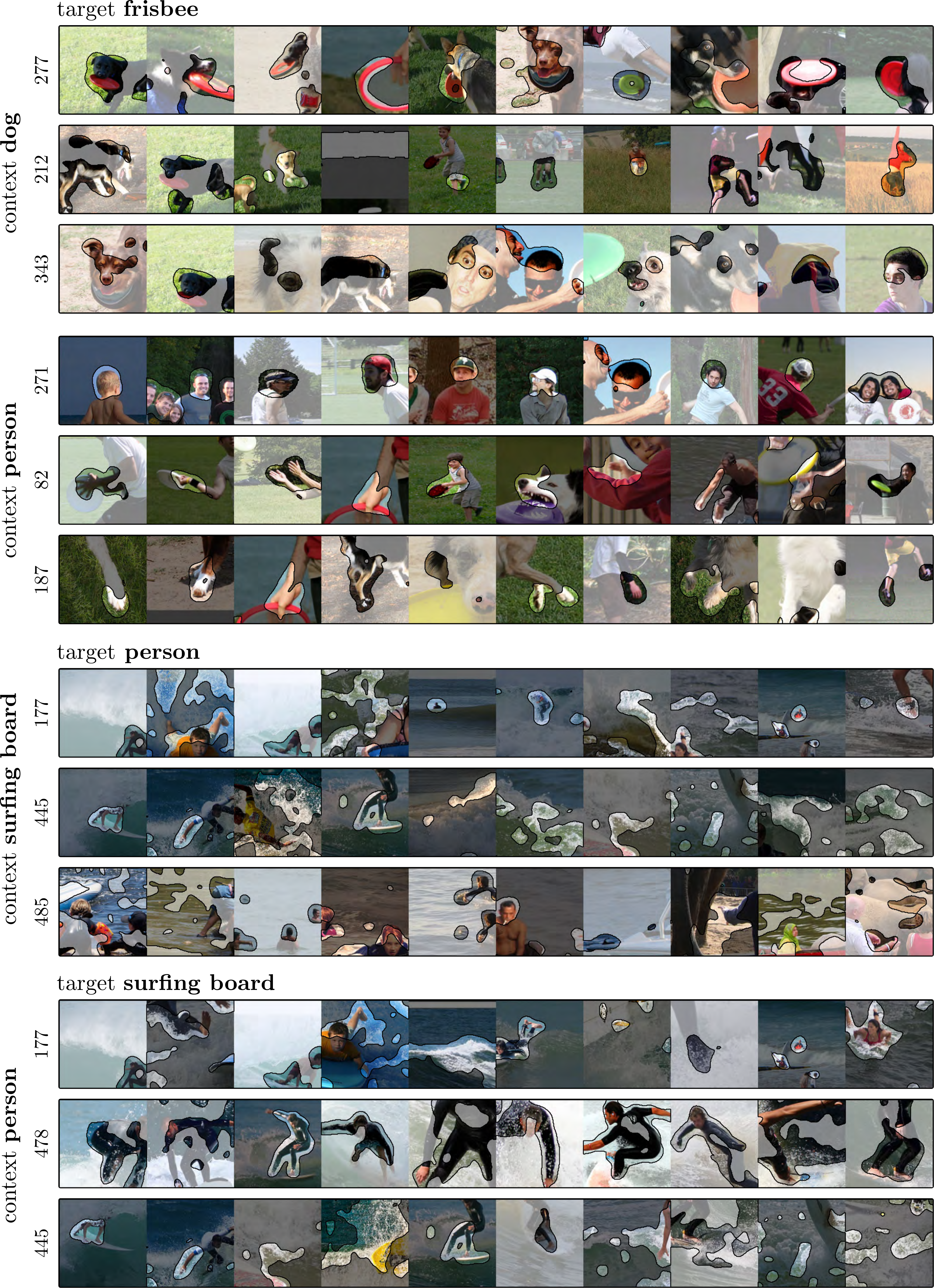}
            \captionof{figure}{Most relevant reference samples for background concepts being flipped for the model interaction experiment in Section~\ref{sec:context}.
            The shown reference samples are conditioned on the corresponding target class displayed above.}
        \label{fig:appendix:context:additional_examples}
            \end{figure*}
    
    In Figure~\ref{fig:appendix:context:additional_examples} we show the identified background concepts in detail with reference samples conditioned on the respective target.
    All concepts correspond to layer \texttt{ERBlock\_5.0.rbr\_dense} of the YOLOv6 model.
    Please note,
    that concepts are flipped in the corresponding order as in Figure~\ref{fig:appendix:context:additional_examples} from top to bottom.
    
    \begin{figure*} 
        \centering
            \includegraphics[width=1\linewidth]{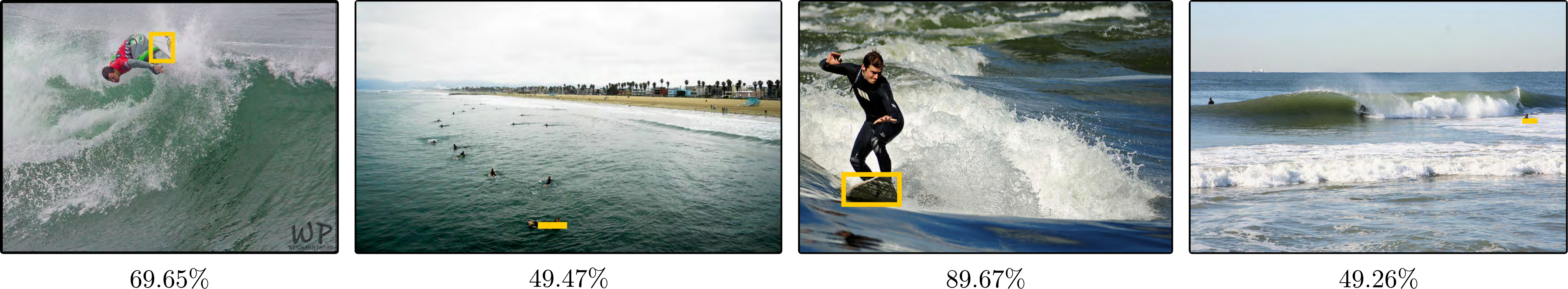}
            \captionof{figure}{Flipping three background concepts (177, 478, 445) of layer \texttt{ERBlock\_5.0.rbr\_dense} of the YOLOv6 model leads to missed surfing board predictions in the shown samples. Initial confidence scores are given below each image.}
        \label{fig:appendix:context:surfing_board}
            \end{figure*}

    Interestingly,
    the perturbation of three concepts of in total 512 leads to missed surfing board predictions.
    In Figure~\ref{fig:appendix:context:surfing_board},
    we show four such examples with initial prediction confidence scores given.

\end{document}